\pdfoutput=1

\documentclass[11pt]{article}

\usepackage{acl}

\usepackage{times}
\usepackage{latexsym}
\usepackage{url}
\usepackage{hyperref}
\usepackage{enumitem}
\usepackage{booktabs}
\usepackage{multirow}
\usepackage{array}
\usepackage{graphicx}
\usepackage{amsmath}
\usepackage{amsfonts}
\usepackage{subcaption}
\usepackage{makecell}
\usepackage{color,soul}
\usepackage{algorithm}
\usepackage[noend]{algpseudocode}
\usepackage{xcolor,colortbl}
\usepackage{xspace}

\definecolor{green}{rgb}{0.1,0.1,0.1}
\definecolor{gitgreen}{HTML}{006400}
\definecolor{chocolate}{HTML}{D2691E}
\definecolor{maroon}{HTML}{A00000}
\definecolor{indigo}{HTML}{4B0082}
\definecolor{green}{HTML}{008000}
\definecolor{red}{HTML}{e41a1c}

\usepackage{amssymb}
\usepackage{pifont}
\newcommand{\cmark}{{\protect\color{maroon} \ding{51}}}
\newcommand{\xmark}{\ding{55}}

\usepackage[T1]{fontenc}

\usepackage[utf8]{inputenc}

\usepackage{microtype}

\newcommand{\ours}{MetaICL}
\newcommand{\ourslong}{\textbf{Meta}-training for \textbf{I}n-\textbf{C}ontext \textbf{L}earning}
\newcommand{\main}{HR$\rightarrow$LR}

\newcommand{\code}{
    \href{https://github.com/facebookresearch/MetaICL}{\nolinkurl{github.com/facebookresearch/MetaICL}}
}

\title{\ours: Learning to Learn In Context}

\newcommand{\affilsup}[1]{\rlap{\textsuperscript{\normalfont#1}}}

\author{
    Sewon Min\affilsup{1,2} \quad 
    ~~~Mike Lewis\affilsup{2} \quad 
    ~~Luke Zettlemoyer\affilsup{1,2} \quad
    ~~~Hannaneh Hajishirzi\affilsup{1,3} 
    \\
    $^1$University of Washington \qquad
    $^2$Meta AI \qquad
    $^3$Allen Institute for AI \\
    \texttt{\{sewon,lsz,hannaneh\}@cs.washington.edu} \qquad 
    \texttt{mikelewis@fb.com} \\
}

\begin{document}
\maketitle
\begin{abstract}
We introduce \ours\ (\ourslong), a new meta-training framework for few-shot learning where a pretrained language model is tuned to do in-context learning on a large set of training tasks.
This meta-training enables the model to more effectively learn a new task in context at test time, by simply conditioning on a few training examples with no parameter updates or task-specific templates.
We experiment on a large, diverse collection of tasks consisting of 142 NLP datasets including classification, question answering, natural language inference, paraphrase detection and more, across seven different meta-training/target splits.
\ours\ outperforms a range of baselines including in-context learning without meta-training and multi-task learning followed by zero-shot transfer.
We find that the gains are particularly significant for target tasks that have domain shifts from the meta-training tasks, and that using a diverse set of the meta-training tasks is key to improvements.
We also show that \ours\ approaches (and sometimes beats) the performance of models fully finetuned on the target task, and outperforms much bigger models with nearly 8x parameters.
Finally, we show that \ours\ is complementary to human-written instructions, and the best performance can be achieved by combining both approaches.
\end{abstract}

\section{Introduction}\label{sec:intro}
Large language models (LMs) have recently been shown to be able to do {\em in-context learning}~\citep{brown2020language}, where they learn a new task simply by conditioning on a few training examples and predicting which tokens best complete a test input. 
This type of learning is attractive because the model learns a new task through inference alone, without any parameter updates.
However, performance significantly lags behind supervised finetuning, results are often high variance~\citep{zhao2021calibrate,perez2021true}, and it can be difficult to engineer the templates that convert existing tasks to this format. 

In this paper, we address these challenges by introducing \ours: \ourslong. \ours\ tunes a pretrained language model on a large set of tasks to learn how to in-context learn, and is evaluated on strictly new unseen tasks.
Each meta-training example matches the test setup---it includes $k+1$ training examples from one task that will be presented together as a single sequence to the language model, and the output of the final example is used to calculate the cross-entropy training loss.
Simply finetuning the model in this data setup directly leads to better in-context learning---the model learns to recover the semantics of the task from the given examples, as must be done for in-context learning of a new task at test time.
This approach is related to recent work that uses multi-task learning for better zero-shot performance at test time~\citep{ khashabi2020unifiedqa,zhong2021adapting,mishra2022cross,wei2022finetuned,sanh2022multitask}.
However, \ours\ is distinct as it allows learning new tasks from $k$ examples alone, without relying on a task reformatting (e.g., reducing everything to question answering) or task-specific templates (e.g., converting different tasks to a language modeling problem).


We experiment on a large, diverse collection of tasks taken from \citet{ye2021crossfit} and \citet{khashabi2020unifiedqa}, including 142 text classification, question answering, natural language inference and paraphrase detection datasets.
We report seven different settings, all with no overlap between meta-training and target tasks. 
This leads to 52 unique target tasks in total, which is the largest among all recent related work to the best of our knowledge.

Experimental results show that \ours\ consistently outperforms baselines including (1) a variety of LM in-context learning baselines without meta-training~\citep{brown2020language,zhao2021calibrate,holtzman2021surface,min2022noisy}, and (2) multi-task learning followed by zero-shot transfer~\citep{zhong2021adapting,wei2022finetuned,sanh2022multitask}.
Gains over multi-task zero-shot transfer are particularly significant when meta-training tasks and target tasks are dissimilar, e.g. there are large differences in task formats, domains, or required skills. This demonstrates that \ours\ enables the model to recover the semantics of the task in context during inference even when the target does not share similarities with meta-training tasks.
\ours\ often gets close to (and sometimes beats) the performance of models trained with supervised finetuning on the target datasets, and perform as well as models with 8x parameters.
We also perform extensive ablations to identify key ingredients for success of \ours\ such as the number and diversity of meta-training tasks. Finally, we demonstrate \ours\ without any templates is better than recent work using human-written natural instructions, while the best performance is achieved by combining both approaches.
Code and data are publicly released at \code.

\section{Related Work}\label{sec:related}\begin{table*}[t]
    \centering \footnotesize
    \begin{tabular}{lll }
        \toprule
            & Meta-training & Inference \\
        \midrule
            Task & $C$ {\em meta-training} tasks & An unseen {\em target} task \\
        \cmidrule{1-3}
            \multirow{2}{*}{Data given}  &
            \multirow{2}{*}{Training examples $\mathcal{T}_i=\{(x^i_j, y^i_j)\}_{j=1}^{N_i},~\forall i \in [1, C]~~~(N_i \gg k)$}
            
            & Training examples $(x_1,y_1),\cdots,(x_k,y_k)$, \\
            &
            & Test input $x$  \\
        \cmidrule{1-3}
            \multirow{4}{*}{Objective} & For each iteration, & \multirow{4}{*}{$\mathrm{argmax}_{c \in \mathcal{C}}P(c|x_1,y_1,\cdots,x_k,y_k,x)$} \\
            & ~~~1. Sample task $i \in [1, C]$ \\
            & ~~~2. Sample $k+1$ examples from $\mathcal{T}_i$: $(x_1,y_1),\cdots,(x_{k+1},y_{k+1})$ \\
            & ~~~3. Maximize $P(y_{k+1}|x_1,y_1,\cdots,x_k,y_k,x_{k+1})$ \\
        \bottomrule
    \end{tabular}\vspace{-.1em}
    \caption{
        Overview of \ours\ (Section~\ref{sec:method}). \ours\ uses the same in-context learning setup at both meta-training and inference. At meta-training time, $k+1$ examples for a task is sampled, where the last example acts as the test example and the rest $k$ examples act as the training examples. Inference is the same as typical in-context learning where $k$ labeled examples are used to make a prediction for a test input.
    }\label{tab:overview}
\end{table*}

\vspace{.2em}
\paragraph{In-context learning}

\citet{brown2020language} propose to use a language model (LM) conditioned on a concatenation of training examples for few-shot learning with no parameter updates.
It has been further improved by later work~\citep{zhao2021calibrate, holtzman2021surface,min2022noisy}, showing promising results on a variety of tasks.
However, in-context learning with an LM achieves poor performance when the target task is very different from language modeling in nature or the LM is not large enough. Moreover, it can have high variance and poor worst-case accuracy~\citep{perez2021true,lu2021fantastically}.

Our paper is based on the core idea of in-context learning by conditioning on training examples. We show that, by explicitly training on an in-context learning objective, \ours\ achieves substantial improvements even with smaller LMs. 

\vspace{.2em}
\paragraph{Meta-training via multi-task learning}
Our work is broadly inspired by a large body of work in meta-learning~\citep{vilalta2002perspective,finn2017model} and multi-task learning~\citep{evgeniou2004regularized,ruder2017overview}.
Prior work has shown that multi-task learning on a large collection of tasks leads to better performance on a new task, either when tested zero-shot~\citep{khashabi2020unifiedqa,zhong2021adapting,mishra2022cross,wei2022finetuned} or when further finetuned~\citep{aghajanyan2021muppet,ye2021crossfit}.
In particular, the former is closely related to our work, as it eliminates the need for parameter updates on a target task.
However, these zero-shot models are either limited to tasks sharing the same format as training tasks (e.g., a question answering format)~\citep{khashabi2020unifiedqa,zhong2021adapting}, or rely heavily on task-specific templates~\citep{mishra2022cross,wei2022finetuned,sanh2022multitask}
which are difficult to engineer due to high variance in performance from very small changes~\citep{mishra2021reframing}.

In this paper, we propose a meta-training method for better in-context learning that improves few-shot performance. We show that it effectively learns semantics of a new task with no manual effort, significantly outperforming zero-shot transfer methods.\footnote{We show that \ours\ without instructions is still better than zero-shot transfer with instructions, but by using instructions, performance of \ours\ further improves (Section~\ref{subsec:ablations}).}
Furthermore, while \citet{wei2022finetuned} show that meta-training helps only when the model has 68B or more parameters, our experiments demonstrate improvements with a much smaller model (770M).

\citet{chen2022meta}, concurrently to our work, propose meta-training for in-context learning. Our approach differs in a number of ways:
we remove requirements of
human-written templates or instructions, and include more diverse tasks, stronger baselines, and extensive experiments in much larger scale with many meta-training/target splits.


\section{\ours}\label{sec:method}

We introduce \ours: \ourslong.
Table~\ref{tab:overview} provides an overview of the approach.
The key idea is to use a multi-task learning scheme over a large collection of meta-training tasks, in order for the model to learn how to condition on a small set of training examples, recover the {\em semantics} of a task, and predict the output based on it. Following previous literature~\citep{brown2020language}, the training examples are concatenated and provided as an single input to the model, which is feasible for $k$-shot learning (e.g., $k=16$).
At test time, the model is evaluated on an unseen target task that comes with $k$ training examples, and inference directly follows the same data format as in meta-training.

\subsection{Meta-training}\label{subsec:meta-training}
The model is meta-trained on a collection of tasks which we call meta-training tasks.
For every iteration, one meta-training task is sampled, and $k+1$ training examples $(x_1,y_1), \cdots, (x_{k+1},y_{k+1})$ are sampled from the training examples of the chosen task.
We then supervise the model by feeding the concatenation of $x_1, y_1, \cdots, x_k, y_k, x_{k+1}$ to the model as an input and train the model to generate $y_{k+1}$ using a negative log likelihood objective.
This simulates in-context learning at inference where the first $k$ examples serve as training examples and the last $(k+1)$-th example is regarded as the test example.

\subsection{Inference}\label{subsec:inference}
For a new target task, the model is given $k$ training examples $(x_1, y_1), \cdots, (x_k, y_k)$ as well as a test input $x$.
It is also given a set of candidates $\mathcal{C}$ which is either a set of labels (in classification) or answer options (in question answering).
As in meta-training, the model takes a concatenation of $x_1, y_1, \cdots, x_k, y_k, x$ as the input, and compute the conditional probability of each label $c_i \in \mathcal{C}$.
The label with the maximum conditional probability is returned as a prediction.

\subsection{Channel \ours}\label{subsec:channel-mic}
We introduce a noisy channel variant of \ours\ called Channel \ours, following \citet{min2022noisy}.
In the noisy channel model, $P(y|x)$ is reparameterized to
$\frac{P(x|y)P(y)}{P(x)} \propto P(x|y)P(y)$.
We follow \citet{min2022noisy} in using $P(y)=\frac{1}{|\mathcal{C}|}$ and modeling $P(x|y)$ which allows us to use the channel approach by simply flipping $x_i$ and $y_i$.
Specifically, at meta-training time, the model is given a concatenation of $y_1, x_1, \cdots, y_k, x_k, y_{k+1}$ and is trained to generate $x_{k+1}$. At inference, the model computes $\mathrm{argmax}_{c \in \mathcal{C}} P(x|y_1,x_1,\cdots,y_k,x_k,c)$. 

\section{Experimental Setup}\label{sec:exp-setup}\begin{table}[t]
    \centering \footnotesize
    \begin{tabular}{l @{\hspace{-.6em}} r @{\hspace{0.4em}} r  l @{\hspace{-.8em}} r}
        \toprule
            \multicolumn{3}{c}{Meta-train} & \multicolumn{2}{c}{Target} \\
            \cmidrule(lr){1-3} \cmidrule(lr){4-5}
            Setting & \# tasks & \# examples & Setting & \# tasks\\
        \midrule
            HR & 61 & 819,200 & LR & 26 \\
        \cmidrule(lr){1-5}
            Classification & 43 & 384,022 & \multirow{2}{*}{Classification} & \multirow{2}{*}{20} \\
            Non-Classification & 37 & 368,768 & & \\
        \cmidrule(lr){1-5}
            QA & 37 & 486,143 & \multirow{2}{*}{QA} & \multirow{2}{*}{22} \\
            Non-QA & 33 & 521,342 & & \\
        \cmidrule(lr){1-5}
            Non-NLI & 55 & 463,579 & NLI & 8 \\
        \cmidrule(lr){1-5}
            Non-Paraphrase & 59 & 496,106 & Paraphrase & 4 \\
        \bottomrule
    \end{tabular}\vspace{-.1em}
    \caption{Statistics of seven different settings.
    Each row indicates meta-training/target tasks for each setting.
    `\# tasks' in meta-training is equivalent to $C$ in Table~\ref{tab:overview}.
    For all settings, there is no overlap in tasks between meta-training and target.
    `HR' and `LR' indicate high resource and low resource, respectively.
    Datasets and the task ontology are taken from  \textsc{CrossFit}~\citep{ye2021crossfit} and \textsc{UnifiedQA}~\citep{khashabi2020unifiedqa}.
    Full datasets for each split are provided in Appendix~\ref{app:dataset}.
    }\label{tab:data-summary}
\end{table}

\begin{table}[t]
    \centering \footnotesize
    \begin{tabular}{l 
        ccc}
        \toprule
            \multirow{2}{*}{Method} &
            Meta &
            \multicolumn{2}{c}{Target}  \\
            \cmidrule(lr){2-2} \cmidrule(lr){3-4}
            & train & train & \# samples \\
        \midrule
            \multicolumn{4}{l}{\textbf{\em LMs}} \\
            0-shot               & \xmark & \xmark & 0 \\
            PMI 0-shot           & \xmark & \xmark & 0 \\
            Channel 0-shot       & \xmark & \xmark & 0 \\
            In-context           & \xmark & \xmark & $k$ \\
            PMI In-context       & \xmark & \xmark & $k$ \\
            Channel In-context   & \xmark & \xmark & $k$ \\
        \midrule
            \multicolumn{4}{l}{\textbf{\em Meta-trained}} \\
            Multi-task 0-shot       & \cmark & \xmark & 0 \\
            Channel Multi-task 0-shot       & \cmark & \xmark & 0 \\
            \ours\ (Ours)            & \cmark & \xmark & $k$ \\
            Channel \ours\ (Ours)            & \cmark & \xmark & $k$ \\
        \midrule
            \multicolumn{4}{l}{\textbf{\em Fine-tune}} \\
            Fine-tune                  & \xmark & \cmark & $k$ \\
            Fine-tune w/ meta-train    & \cmark & \cmark & $k$ \\
        \bottomrule
    \end{tabular}\vspace{-.3em}
    \caption{Summary of the baselines and \ours.
    `train' indicates whether the model is trained with parameter updates, and `\# samples' indicates the number of training examples used on a target task.
    Our baselines include a range of recently introduced methods~\citep{holtzman2021surface,zhao2021calibrate,min2022noisy,wei2022finetuned} as described in Section~\ref{subsec:baselines}.
    }\label{tab:methods}
\end{table}

\subsection{Datasets}\label{subsec:dataset}
We use a large collection of tasks taken from \textsc{CrossFit}~\citep{ye2021crossfit} and \textsc{UnifiedQA}~\citep{khashabi2020unifiedqa}.
We have 142 unique tasks in total, covering a variety of problems including text classification, question answering (QA), natural language inference (NLI) and paraphrase detection. All tasks are in English.

We experiment with seven distinct settings as shown in Table~\ref{tab:data-summary}, where there is no overlap between the meta-training and target tasks. The number of unique target tasks in total is 52, which is significantly larger than other relevant work~\citep{khashabi2020unifiedqa,zhong2021adapting,mishra2022cross,wei2022finetuned,sanh2022multitask}.
Each target task is either classification or multi-choice, where a set of candidate options ($\mathcal{C}$ in Table~\ref{tab:overview}) is given.

\vspace{.25em}
\noindent \textbf{\main} (High resource to low resource): We experiment with a setting where datasets with 10,000 or more training examples are used as meta-training tasks and the rest are used as target tasks. We think using high resource datasets for meta-training and low resource datasets as targets is a realistic and practical setting for few-shot learning.

\vspace{.25em}
\noindent \textbf{X$\rightarrow$X (X=\{Classification, QA\})}: We experiment with two settings with meta-training and target tasks sharing the task format, although with no overlap in tasks.

\vspace{.25em}
\noindent \textbf{Non-X$\rightarrow$X (X=\{Classification, QA, NLI, Paraphase\})}: Lastly, we experiment with four settings where meta-training tasks do not overlap with target tasks in task format and required capabilities. These settings require the most challenging generalization capacities.

\vspace{.25em}
Each setting has a subset of target tasks with no domain overlap with any meta-training tasks (e.g., finance, poem, climate or  medical).
We report both on all target tasks or on target tasks with no domain overlap only.
Full details of the settings and datasets with citations are provided in Appendix~\ref{app:dataset}.

\subsection{Baselines}\label{subsec:baselines}
We compare \ours\ and Channel \ours\ with a range of baselines, as summarized in Table~\ref{tab:methods}.

\vspace{.2em}
\noindent \textbf{0-shot}: We use a pretrained LM as it is and run zero-shot inference, following \citet{brown2020language}.

\vspace{.2em}
\noindent \textbf{In-context}: We use the pretrained LM as it is and use in-context learning by conditioning on a concatenation of $k$ training examples, following \citet{brown2020language}.

\vspace{.2em}
\noindent \textbf{PMI 0-shot, PMI In-context}: We use the PMI method from \citet{holtzman2021surface,zhao2021calibrate} for 0-shot and In-context learning.

\vspace{.2em}
\noindent \textbf{Channel 0-shot, Channel In-context}: We use the noisy channel model from \citet{min2022noisy} for 0-shot and In-context learning.

\vspace{.2em}
\noindent \textbf{Multi-task 0-shot}: We train the LM on the same meta-training tasks without in-context learning objective, i.e., maximize $P(y|x)$ without $k$ other training examples, and then use zero-shot transfer on a target task. This is equivalent to \ours\ with $k=0$. This is a typical multi-task learning approach from previous work~\citep{khashabi2020unifiedqa,zhong2021adapting,wei2022finetuned}.

\vspace{.2em}
\noindent \textbf{Channel Multi-task 0-shot}: We have a channel variant of Multi-task 0-shot.

\vspace{.2em}
\noindent \textbf{Fine-tune}: We fine-tune the LM on an individual target task. This is not directly comparable to other methods as parameter updates are required for every target task.

\vspace{.2em}
\noindent \textbf{Fine-tune w/ meta-train}: We train the LM on meta-training tasks first and then further fine-tuned it on a target task. This is not directly comparable to other methods for the same reason as above.


\newcommand{\prom}{{\protect\color{green} [P]}}
\newcommand{\hyp}{{\protect\color{maroon} [H]}}

\begin{table}[t]
    \centering \footnotesize
    \setlength{\tabcolsep}{0.3em}
    \begin{tabular}{ll}
        \toprule
            \multicolumn{2}{l}{\prom: Time Warner is the world’s largest media and Internet} \\ \multicolumn{2}{l}{company.} \\
            \multicolumn{2}{l}{\hyp: Time Warner is the world’s largest company.} \\
            \multicolumn{2}{l}{Labels: \texttt{entailment}, \texttt{not\_entailment}} \\
        \midrule
           \multicolumn{2}{l}{\textbf{\em \citet{holtzman2021surface}}} \\
           Input    & \prom\ question: \hyp\ true or false? answer: \\
           Output   & $\{$true, false$\}$
           \\
        \cmidrule{1-2}
           \multicolumn{2}{l}{\textbf{\em \citet{wei2022finetuned}}} \\
           Input & \prom\ Based on the paragraph above, can we \\
           & conclude that \hyp? \\
           Output & $\{$yes, no$\}$
           \\
         \cmidrule{1-2}
           \multicolumn{2}{l}{\textbf{\em Ours}} \\
           Input & \prom\ \hyp\ \\
           Output & $\{$entailment, not\_entailment$\}$ \\
        \bottomrule
    \end{tabular}
    \caption{Example input-output pairs for an NLI task.
    We show human-authored templates taken from prior work as references.
    }\label{tab:templates}
\end{table}

\newcolumntype{C}[1]{>{\PreserveBackslash\centering}p{#1}}

\begin{table*}[t]
    \centering \footnotesize
    \begin{tabular}{
        l @{\hspace{2em}}
        ccccccc
        }
        \toprule
            Method &
            \main &
            \makecell[c]{Class \\ $\rightarrow$Class} & \makecell[c]{non-Class \\ $\rightarrow$Class} &
            \makecell[c]{QA \\ $\rightarrow$QA} &
            \makecell[c]{non-QA \\ $\rightarrow$QA} &
            \makecell[c]{non-NLI \\ $\rightarrow$NLI} &
            \makecell[c]{non-Para \\ $\rightarrow$Para} \\
        \midrule
            \multicolumn{8}{c}{\em All target tasks} \\
            0-shot & 34.8 & 34.2 & 34.2 & 40.2 & 40.2  & 25.5 & 34.2 \\
            PMI 0-shot & 35.1 & 33.8 & 33.8 & 40.2 & 40.2 & 27.9 & 39.2 \\
            Channel 0-shot & 36.5 & 37.3 & 37.3 & 38.7 & 38.7 & 33.9 & 39.5 \\
            In-context & 38.2/35.3 & 37.4/33.9 & 37.4/33.9 & 40.1/38.7 & 40.1/38.7 & 34.0/28.3 & 33.7/33.1 \\
            PMI In-context & 39.2/33.7 & 38.8/30.0 & 38.8/30.0 & 40.3/38.8 & 40.3/38.8 & 33.0/28.0 & 38.6/33.4 \\
            Channel In-context & 43.1/38.5 & 46.3/40.3 & 46.3/40.3 & 40.8/38.1 & 40.8/38.1 & 39.9/34.8 & 45.4/40.9 \\
        \cmidrule{1-8}
            Multi-task 0-shot & 35.6 & 37.3 & 36.8 & 45.7 & 36.0 & 40.7 & 30.6 \\
            Channel Multi-task 0-shot & 38.8 & 40.9 & 42.2 & 42.1 & 36.4 & 36.8 & 35.1 \\
            \ours & 43.3/41.7 & 43.4/39.9 & 38.1/31.8 & \textbf{46.0}/44.8 & 38.5/36.8 & 49.0/44.8 & 33.1/33.1 \\
            Channel \ours & \textbf{49.1}/46.8 & \textbf{50.7}/48.0 & \textbf{50.6}/48.1 & 44.9/43.5 & \textbf{41.9}/40.5 & \textbf{54.6}/51.9 & \textbf{52.2}/50.3 \\
        \cmidrule{1-8}
            Fine-tune              & 46.4/40.0 & 50.7/44.0 & 50.7/44.0 & 41.8/39.1 & 41.8/39.1 & 44.3/32.8 & 54.7/48.9 \\
            Fine-tune w/ meta-train& 52.0/47.9 & 53.5/48.5 & 51.2/44.9 & 46.7/44.5 & 41.8/39.5 & 57.0/44.6 & 53.7/46.9 \\
        \toprule
            \multicolumn{8}{c}{\em Target tasks in unseen domains} \\
            0-shot & 32.6 & 32.6 & 32.6 & 45.9 & 45.9 & 33.4 & 38.3 \\
            PMI 0-shot & 28.9 & 28.9 & 28.9 & 44.4 & 44.4 & 33.4 & 32.9 \\
            Channel 0-shot & 29.1 & 29.1 & 29.1 & 41.6 & 41.6 & 33.1 & 32.6 \\
            In-context & 30.6/27.5 & 30.6/27.5 & 30.6/27.5 & 45.6/44.7 & 45.6/44.7 & 52.0/41.3 & 35.8/34.1 \\
            PMI In-context  & 34.9/27.7 & 34.9/27.7 & 34.9/27.7 & 45.4/44.7 & 45.4/44.7 & 47.8/35.2 & 38.5/33.3 \\
            Channel In-context & 39.6/33.6 & 39.6/33.6 & 39.6/33.6 & 44.7/40.6 & 44.7/40.6 & 40.4/35.7 & 44.1/36.8 \\
        \cmidrule{1-8}
            Multi-task 0-shot & 35.4 & 28.0 & 28.6 & \textbf{71.2} & 40.3 & 33.5 & 35.0 \\
            Channel Multi-task 0-shot & 36.3 & 31.1 & 34.3 & 54.4 & 39.4 & 50.8 & 34.1 \\
            \ours & 35.3/32.7 & 32.3/29.3 & 28.1/25.1 & 69.9/68.1 & \textbf{48.3}/47.2 & \textbf{80.1}/77.2 & 34.0/34.0 \\
            Channel \ours & \textbf{47.7}/44.7 & \textbf{41.9}/37.8 & \textbf{48.0}/45.2 & 57.9/56.6 & 47.2/45.0 & 62.0/57.3 & \textbf{51.0}/49.9 \\
        \cmidrule{1-8}
            Fine-tune              & 44.9/37.6 & 44.9/37.6 & 44.9/37.6 & 43.6/39.1 & 43.6/39.1 & 56.3/33.4 & 56.6/51.6 \\
            Fine-tune w/ meta-train& 53.3/43.2 & 53.2/43.7 & 46.1/36.9 & 67.9/66.2 & 44.5/42.8 & 71.8/58.2 & 65.6/61.4 \\
        \bottomrule
    \end{tabular}\vspace{-.3em}
    \caption{Main results, using GPT-2 Large.
    Two numbers indicate the average and the worst-case performance over different seeds used for $k$ target training examples.
    \textbf{Bold} indicates the best average result except results from fine-tuned models that are not comparable.
    `Class' indicates `Classification'.
    }\label{tab:main-result}
\end{table*}

\subsection{Evaluation}\label{subsec:evaluation}
We use Macro-F1\footnote{More suitable than accuracy for imbalanced classification.} and Accuracy as evaluation metrics for classification tasks and non-classification tasks, respectively. 

For a target task, we use $k=16$ training examples, sampled uniformly at random.
We relax the assumption of perfect balance between labels on $k$ training examples, following \citet{min2022noisy}.
Because in-context learning is known to have high variance~\citep{zhao2021calibrate,perez2021true,lu2021fantastically}, we use 5 different sets of $k$ training examples.
We first compute the average and the worst-case performance over seeds for every target task, and then report the macro-average of them over all target tasks.

\subsection{Experiment Details}\label{subsec:impl-details}
As a base LM, we use GPT-2 Large ~\citep{radford2019language} which consists of 770M parameters.\footnote{Appendix~\ref{app:abl_lm_size} reports performance for other LM sizes.}
For baselines without meta-training (raw LMs), we also compare with GPT-J~\citep{wang2021gpt}, which is the largest public causal LM at the time of writing, consisting of 6B parameters.

\begin{table*}[t]
    \centering \footnotesize
    \begin{tabular}{
        l @{\hspace{1.5em}}
        ccccccc
        }
        \toprule
            Method &
            \main\ &
            \makecell[c]{Class \\ $\rightarrow$Class} & \makecell[c]{non-Class \\ $\rightarrow$Class} &
            \makecell[c]{QA \\ $\rightarrow$QA} &
            \makecell[c]{non-QA \\ $\rightarrow$QA} &
            \makecell[c]{non-NLI \\ $\rightarrow$NLI} &
            \makecell[c]{non-Para \\ $\rightarrow$Para} \\
        \midrule
            \multicolumn{8}{c}{\em All target tasks} \\
            Channel In-context & 43.1/38.5 & 46.3/40.3 & 46.3/40.3 & 40.8/38.1 & 40.8/38.1 & 39.9/34.8 & 45.4/40.9 \\
            \ours & 43.3/41.7 & 43.4/39.9 & 38.1/31.8 & 46.0/44.8 & 38.5/36.8 & 49.0/44.8 & 33.1/33.1 \\
            Channel \ours & \textbf{49.1}/46.8 & 50.7/48.0 & 50.6/48.1 & 44.9/43.5 & 42.1/40.8 & \textbf{54.6}/51.9 & \textbf{52.2}/50.3 \\
	    \cmidrule{1-8}
	        GPT-J Channel In-context & 48.6/44.4 & \textbf{51.5}/47.0 & \textbf{51.5}/47.0 & \textbf{47.0}/45.2 & \textbf{47.0}/45.2 & 47.2/41.7 & 51.0/47.5 \\
        \toprule
            \multicolumn{8}{c}{\em Target tasks in unseen domains} \\
            Channel In-context & 39.6/33.6 & 39.6/33.6 & 39.6/33.6 & 44.7/40.6 & 44.7/40.6 & 40.4/35.7 & 44.1/36.8 \\
            \ours & 35.3/32.7 & 32.3/29.3 & 28.1/25.1 & \textbf{69.9}/68.1 & 48.3/47.2 & \textbf{80.1}/77.2 & 34.0/34.0 \\
            Channel \ours & \textbf{47.7}/44.7 & 41.9/37.8 & \textbf{48.0}/45.2 & 57.9/56.6 & 47.2/45.0 & 62.0/57.3 & 51.0/49.9 \\
	    \cmidrule{1-8}
	        GPT-J Channel In-context & 42.8/38.4 & \textbf{42.8}/38.4 & 42.8/38.4 & 55.7/54.4 & \textbf{55.7}/54.4 & 51.1/40.4 & \textbf{52.0}/46.5 \\
        \bottomrule
    \end{tabular}\vspace{-.3em}
    \caption{
    Comparison between raw LM in-context learning (based on GPT-2 Large and GPT-J) and \ours\ (based on GPT-2 Large). GPT-2 Large used unless otherwise specified.
    Two numbers indicate the average and the worst-case performance over different seeds used for $k$ target training examples.
    For raw LM baselines, Channel In-context is reported because it is the best raw LM baseline overall across the settings; full results based on GPT-J are provided in Appendix~\ref{app:gpt-j-result}.
    }\label{tab:comparison-gptj}
\end{table*}

\vspace{-.25em}
\paragraph{Elimination of templates}
Prior work uses human-authored templates to transform the input-output pair to a natural language sentence~\citep{zhong2021adapting,mishra2022cross,wei2022finetuned,chen2022meta}.
They require expensive manual effort (as 136 different templates are required for 136 tasks in this paper) and cause unstable model performance due to many different ways of writing~\citep{mishra2021reframing}.
We eliminate templates, using the given input (or a concatenation of inputs if there are multiple) and label words provided in the original datasets.\footnote{In our preliminary experiments, we explored templates taken from prior work, but found that they do not consistently improve few-shot performance, even when they do improve zero-shot performance.}
A comparison of input-output schemes from prior work and our approach is shown in Table~\ref{tab:templates}.

\paragraph{Training details}
All implementation is done in PyTorch~\citep{paszke2019pytorch} and Transformers~\citep{wolf-etal-2020-transformers}. For meta-training, we use up to 16,384 training examples per task. We use a batch size of $8$, learning rate of $1 \times 10^{-5}$ and a sequence length of $1024$. For multi-task 0-shot baselines (the baselines with no in-context learning), we use a sequence length of $256$. We train the model for $30,000$ steps.\footnote{We also explored training longer, but it did not improve performance.} To save memory during meta-training, we use an 8-bit approximation~\citep{dettmers20228} of an Adam optimizer~\citep{kingma2015adam} and mixed precision~\citep{micikevicius2017mixed}. Training was done for 4.5 hours with eight 32GB GPUs. This is drastically more efficient than recent prior work, e.g., 270 hours of a 512GB TPU in \citet{sanh2022multitask}.

\vspace{.2em}
More details about preprocessing and training can be found in Appendix~\ref{app:impl-details}.

\section{Experimental Results}\label{sec:exp-result}


\subsection{Main Results}\label{subsec:main-results}

Table~\ref{tab:main-result} reports the full results using GPT-2 Large, where we compute the average and the worst-case performance of every target task and report the macro-average over them.
The top and the bottom respectively evaluate on all target tasks and target tasks in unseen domains only.

\vspace{-.2em}
\paragraph{Our baselines are strong}
We first discuss the results of ours baselines.
Among raw LMs without meta-training (the first six rows of Table~\ref{tab:main-result}), we observe that channel in-context baselines are the most competitive, consistent with findings from \citet{min2022noisy}.
We then find that Multi-task 0-shot baselines do not outperform the best raw LM baseline in most settings, despite being supervised on a large set of meta-training tasks.
This somewhat contradicts findings from \citet{wei2022finetuned,sanh2022multitask}.
This is likely for two reasons.
First, our models are much smaller than theirs (770M vs. 11B--137B); in fact, \citet{wei2022finetuned} reports Multi-task 0-shot starts to be better than raw LMs only when the model size is 68B or larger.
Second, we compare with much stronger channel baselines which they did not;
Multi-task 0-shot outperforms non-channel LM baselines but not channel LM baselines.

\vspace{-.2em}
\paragraph{\ours\ outperforms baselines}
\ours\ and Channel \ours\ consistently outperform a range of strong baselines.
In particular, Channel \ours\ achieves the best performance in 6 out of 7 settings.
Gains are particularly significant in the \main, non-NLI$\rightarrow$NLI and non-Para$\rightarrow$Para settings (6--15\% absolute).
This is noteworthy because \main\ targets the common low-resource case where new tasks have very few labeled examples, and the other two represent large data distribution shifts where the test tasks are relatively different from the meta-training tasks. This demonstrates that \ours\ can infer the semantics of new tasks in context even when there are no closely related training tasks.

While \ours\ significantly outperforms baselines in most settings, it only marginally outperforms Multi-task 0-shot in the QA$\rightarrow$QA setting, as an exception.
This is likely because the meta-training and target tasks are relatively similar, allowing the Multi-task 0-shot baseline to achieve very strong performance.
Nonetheless, performance of Multi-task 0-shot in QA significantly drops when the model is trained on non-QA tasks, while performance of \ours\ drops substantially less.

\begin{figure}[!t]
\resizebox{\columnwidth}{!}{
    \includegraphics[trim={0.2cm 0.2cm 0.2cm 0.2cm},clip]{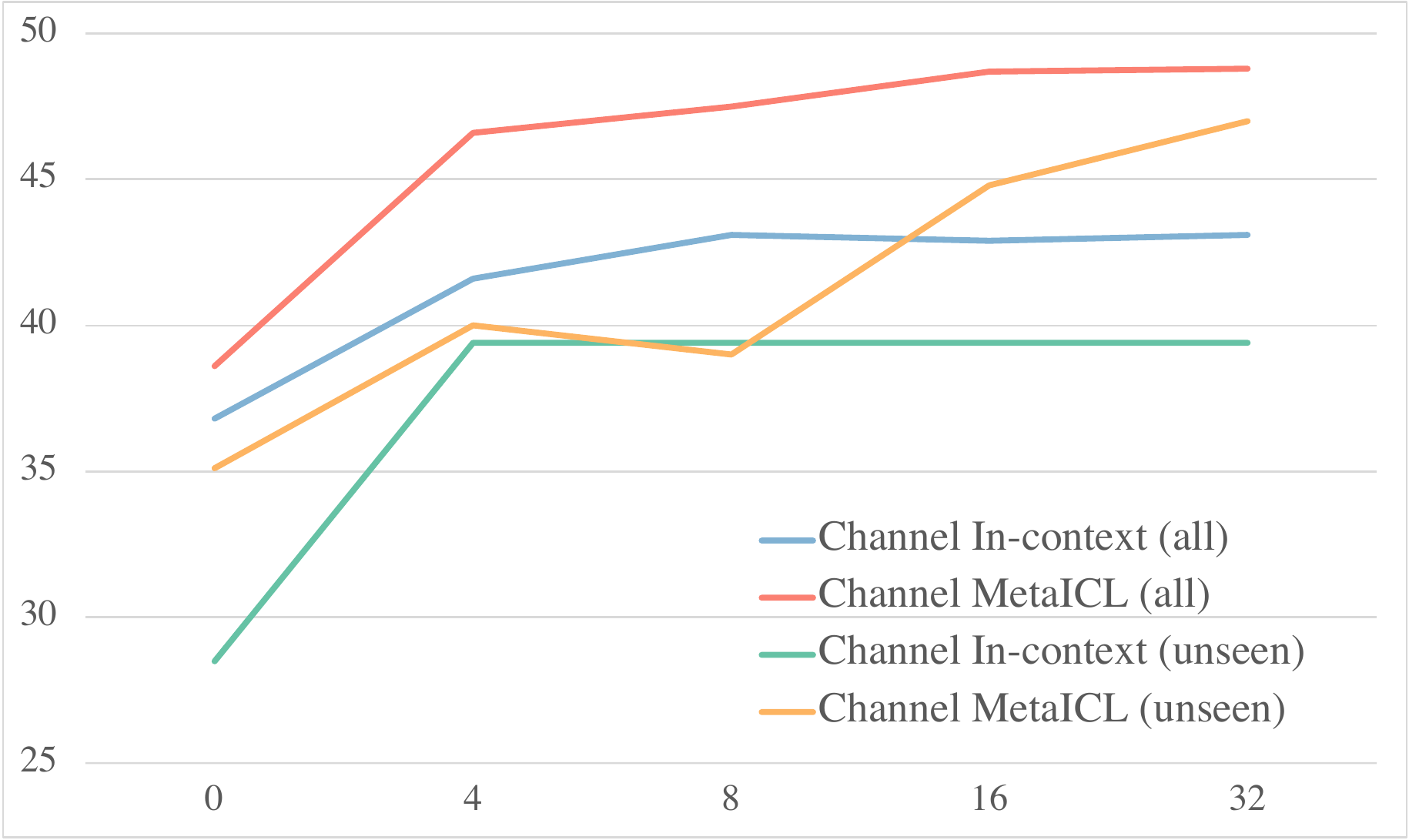}
}
\caption{Ablation on the number of training examples ($k$) in the \main\ setting. $k=0$ is equivalent to the zero-shot methods.}\label{fig:ablate_k}
\end{figure}

\vspace{-.2em}
\paragraph{Gains are larger on unseen domains}
Gains over Multi-task 0-shot are more significant on target tasks in unseen domains.
In particular, Multi-task 0-shot is generally less competitive compared to raw LM baselines, likely because they require more challenging generalization.
\ours\ suffers less from this problem and is consistently better or comparable to raw LM baselines across all settings.

\vspace{-.2em}
\paragraph{Comparison to fine-tuning}
\ours\ matches or sometimes even outperforms fine-tuned models without meta-training.
This is a promising signal, given that no prior work has shown models with no parameter updates on the target can match or outperform supervised models.
Nonetheless, fine-tuning with meta-training exceeds both \ours\ and fine-tuning without meta-training, because meta-training helps in supervised learning as it does in in-context learning.
This indicates that there is still room for improvement in methods that allow learning without parameter updates .

\vspace{-.2em}
\paragraph{Comparison to GPT-J}
In Table~\ref{tab:comparison-gptj}, we compare GPT-2 Large based models with raw LM baselines based on GPT-J which consists of 6B parameters. \ours, despite being 8x smaller, outperforms or matches GPT-J baselines. 


\subsection{Ablations}\label{subsec:ablations}

\paragraph{Varying number of training examples}
We vary the number of training examples ($k$) from 0, 4, 8, 16 to 32. In-context learning with $k=0$ is equivalent to the zero-shot method. Results are shown in Figure~\ref{fig:ablate_k}. Increasing $k$ generally helps across all models, and Channel MetaICL outperforms the raw in-context learning over all values of $k$.
We additionally find that the performance tends to saturate when $k$ is closer to $16$, likely because the sequence length limit of the language model makes it hard to encode many training examples.

\begin{figure}[!t]
\resizebox{\columnwidth}{!}{
    \includegraphics[trim={0.2cm 0.2cm 0.2cm 0.2cm},clip]{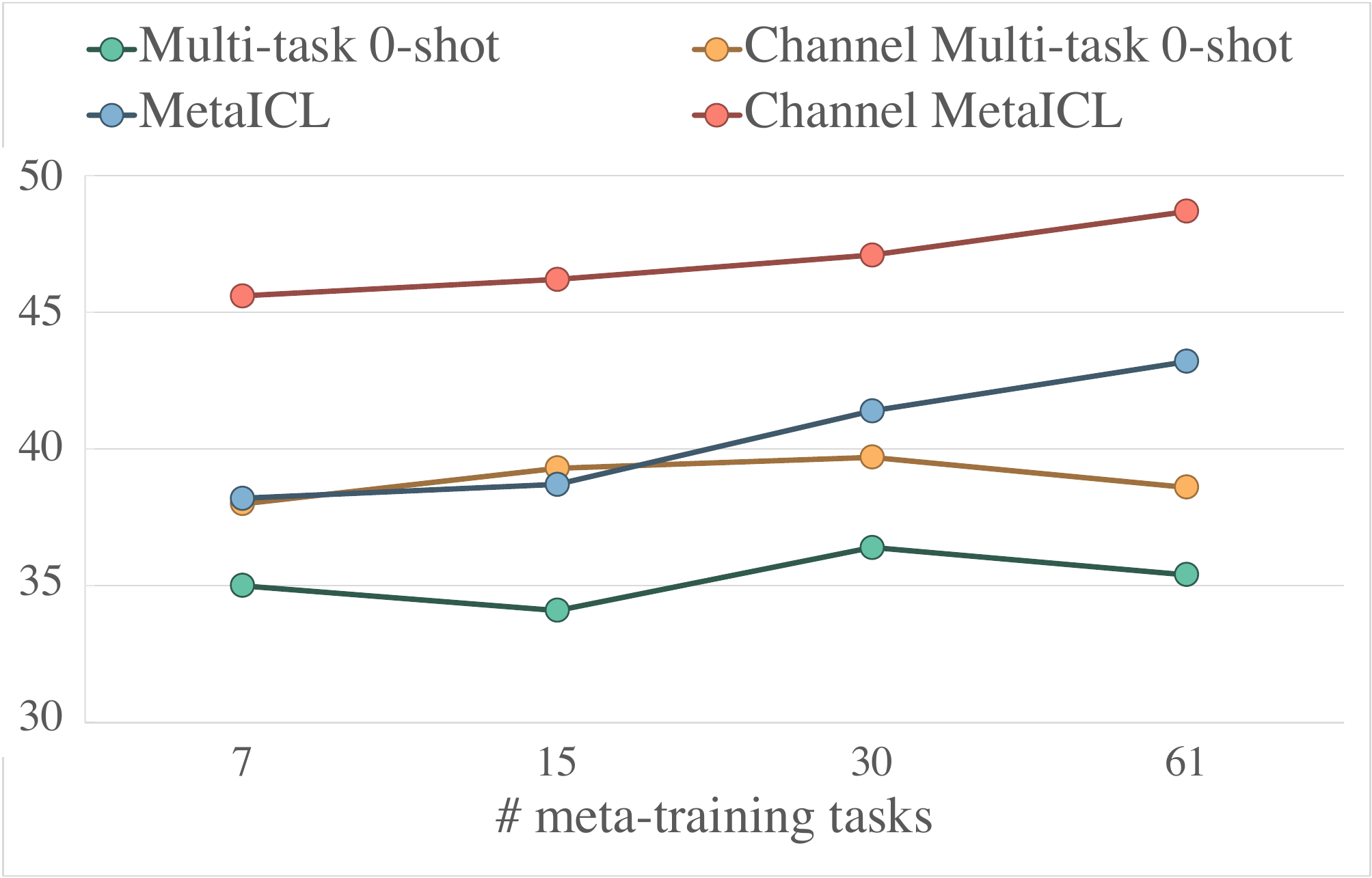}
}
\resizebox{\columnwidth}{!}{
    \includegraphics[trim={0.2cm 0.2cm 0.2cm 0.2cm},clip]{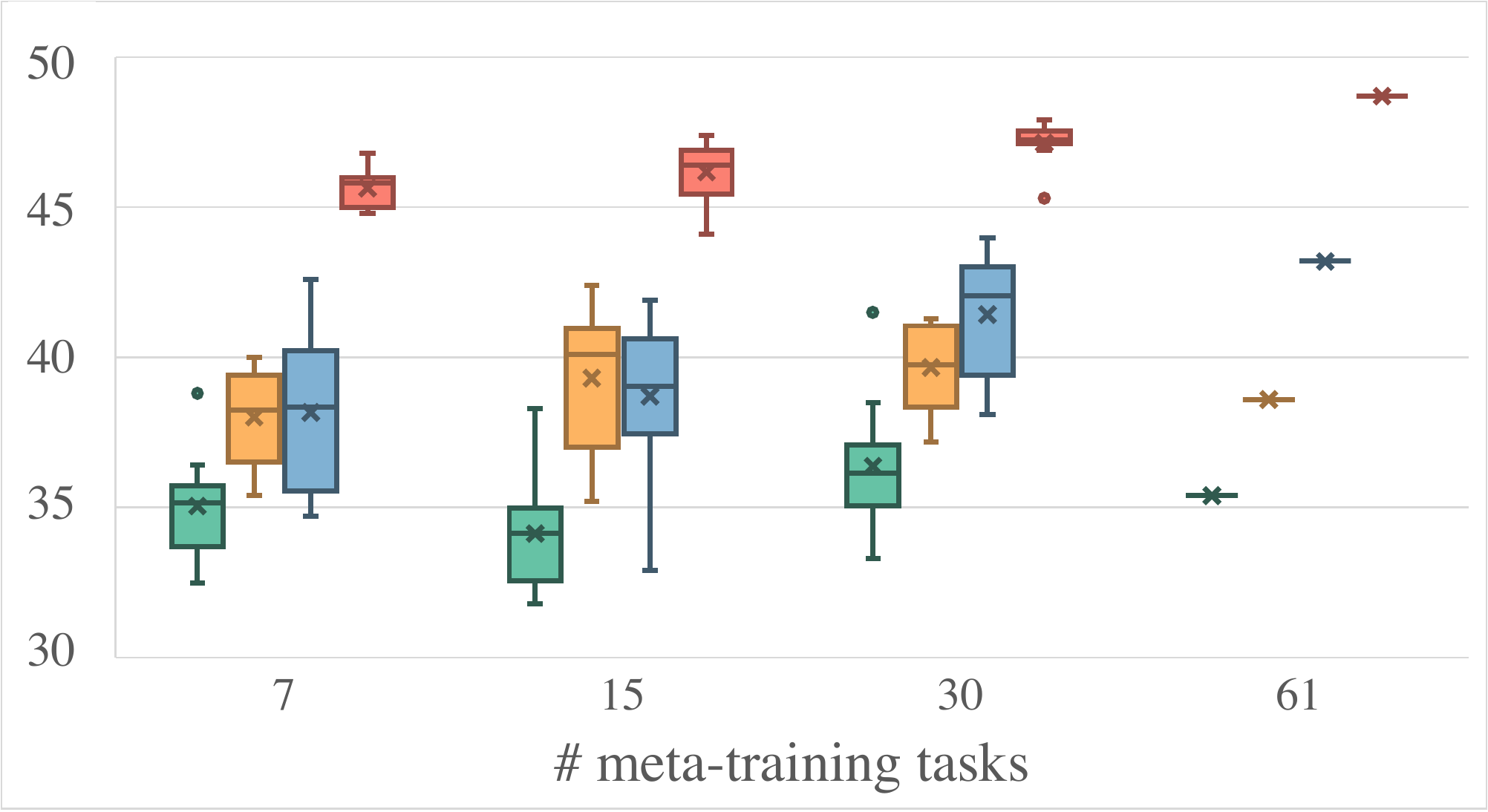}
}
\caption{Ablation on the number of meta-training tasks ($\{7,15,30,61\}$).
The graph of the average (top) and the box chart (bottom) over different meta-training sets using 10 different random seeds (except for $61$).
}\label{fig:ablate_size}
\end{figure}

\vspace{-.2em}
\paragraph{Number of meta-training tasks}
To see the impact of the number of meta-training tasks,
we subsample $\{7,15,30\}$ meta-training tasks out of 61 in the \main\ setting. For each, we use ten different random seeds to additionally see the impact of the choice of meta-training tasks.

Figure~\ref{fig:ablate_size} reports the results.
On average, performance generally increases as the number of tasks increase, which is consistent with results in \citet{mishra2022cross,wei2022finetuned}.
Across different numbers of meta-training tasks, Channel \ours\ consistently outperforms other models.
Nonetheless, there is nonnegligible variance across different choices of meta-training (the bottom of Figure~\ref{fig:ablate_size}), indicating that a choice of meta-training gives substantial impact in performance.

\vspace{-.2em}
\paragraph{Diversity in meta-training tasks}
We hypothesize that the diversity in meta-training tasks may impact performance of \ours. 
To verify this hypothesis, we create two settings by subsampling 13 out of 61 meta-training datasets in the \main\ setting.
One setting is diverse in their task formats and required capacities: QA, NLI, relation extraction, sentiment analysis, topic classification, hate speech detection and more.
The other setting is less diverse, including tasks related to sentiment analysis, topic classification and hate speech detection only.
A full list of datasets is reported in Appendix~\ref{app:dataset}. Using these two settings, we compare multi-task zero-shot transfer baselines and \ours.

Results are reported in Table~\ref{tab:ablate_diversity}.
We find that \ours\ with a diverse set outperforms \ours\ with a non-diverse set by a substantial margin.
This shows that diversity among meta-training tasks is one of substantial factors for the success of \ours.
%

In Appendix~\ref{app:which-tasks-helpful}, we include ablations that provide more insights on the choice of meta-training tasks, such as (1) high quality data with diverse domains tend to help (e.g., GLUE family~\citep{wang2018glue}) and (2) adversarially collected data tends to be unhelpful. However, more systematic studies on how to choose the best meta-training tasks and how they relate to particular target tasks should be done, which we leave for future work.

\begin{table}[!t]
    \centering \footnotesize
    \begin{tabular}{
        l @{\hspace{2em}} cc
        }
        \toprule
            Method &
            Diverse & No Diverse \\
        \midrule
            0-shot & \multicolumn{2}{c}{34.9} \\
            PMI 0-shot & \multicolumn{2}{c}{34.8} \\
            Channel 0-shot & \multicolumn{2}{c}{36.8} \\
            In-context & \multicolumn{2}{c}{38.2/35.4} \\
            PMI In-context & \multicolumn{2}{c}{38.9/33.3} \\
            Channel In-context & \multicolumn{2}{c}{42.9/38.5} \\
        \cmidrule{1-3}
            Multi-task 0-shot           & 35.2 & 29.9 \\
            Channel Multi-task 0-shot   & 41.6 & 38.3 \\
            \ours               & 45.6/43.4 & 38.8/35.4 \\
            Channel \ours       & \textbf{47.2}/44.7 & 45.3/42.6 \\
        \bottomrule
    \end{tabular}\vspace{-.3em}
    \caption{Ablation on the diversity of meta-training tasks in the \main\ setting. For both settings, the number of meta-training tasks is 13, and the number of target tasks is 26 as in the original \main\ setting.
    A full list of meta-training tasks is shown in Appendix~\ref{app:dataset}.
    }\label{tab:ablate_diversity}
\end{table}

\vspace{-.2em}
\paragraph{Are instructions necessary?}
Most recent work has used human-written natural instructions for zero- or few-shot learning~\citep{mishra2022cross, wei2022finetuned,sanh2022multitask}.
While we argue for not using instructions to avoid manual engineering and high variance, we also ask: {\em are instructions still useful with \ours?} On one hand, learning to condition on $k$ examples may remove the necessity of instructions. On the other hand, instructions may still be complementary and provide the model with extra useful infomration.

We aim to answer this question by using 32 meta-training tasks and 12 target tasks from the \main\ setting for which human-written instructions are available in \citet{sanh2022multitask}.\footnote{\url{github.com/bigscience-workshop/promptsource}}
We have two variants: (a) using one instruction per meta-training task, and (b) using all available instructions including 267 instructions in total (8.3 per meta-training task) which \citet{sanh2022multitask} found to be better than (a).
We then compare \ours\ and a range of baselines with and without instructions.

\begin{table}[t]
    \centering \footnotesize
    \begin{tabular}{
        l @{\hspace{0.6em}} c @{\hspace{1em}} c @{\hspace{1em}} c
        }
        \toprule
            Method &
            w/o Instruct & \multicolumn{2}{c}{w/ Instruct} \\
        \midrule
            \# instruct/task & 0 & 1 & 8.3 \\
        \midrule
            0-shot          & 33.3 & \multicolumn{2}{c}{34.2}  \\
            PMI 0-shot      & 34.6 & \multicolumn{2}{c}{27.8}  \\
            Channel 0-shot  & 32.5 & \multicolumn{2}{c}{30.6}  \\
            In-context      & 34.5/31.5 & \multicolumn{2}{c}{45.2/42.3}  \\
            PMI In-context  & 37.7/32.7 & \multicolumn{2}{c}{41.9/37.6}  \\
            Channel In-context & 39.0/35.4 & \multicolumn{2}{c}{39.6/35.3}  \\
        \cmidrule{1-4}
            MT 0-shot           & 35.7 & 32.6 & 37.1 \\
            Channel MT 0-shot   & 36.7 & 30.6 & 36.0 \\
            \ours               & 40.4/37.7 & 42.6/41.0 & 43.2/41.0 \\
            Channel \ours       & 42.2/40.0 & 45.3/43.9 & \textbf{46.9}/44.2 \\
        \bottomrule
    \end{tabular}\vspace{-.3em}
    \caption{Ablation on the impact of {\em natural instructions}.
    `w/ Instruct' uses instructions from \citet{sanh2022multitask}, either one per meta-training task or all available ones;
    `w/o Instruct' does not use instructions, as in all of our other experiments.
    `\# instruct/task' indicates the number of instructions per meta-training task on average.
    `MT 0-shot' indicates `Multi-task 0-shot' baselines.
    Both settings have the same meta-training and target tasks, 
    32 and 12, respectively.
    A full list of tasks is shown in Appendix~\ref{app:dataset}.
    }\label{tab:ablate_inst}
\end{table}

Results are reported Table~\ref{tab:ablate_inst}.
As in \citet{wei2022finetuned} and \citet{sanh2022multitask}, Multi-task 0-shot outperforms the raw-LM 0-shot baseline.
However, \ours\ with no instructions is better than Multi-task 0-shot with instructions.
Furthermore, \ours\ achieves further improvements when instructions are jointly used, significantly outperforming all baselines.
In fact, 
when increasing the number of instructions per task from 0, 1 to 8.3,
performance of \ours\ improves much more than performance of Multi-task 0-shot does.
To summarize, (1) learning to in-context learn (\ours) outperforms learning to learn from instructions; (2) \ours\ and using instructions are largely complementary, and (3) \ours\ actually benefits more from using instructions than Multi-task 0-shot does.

Importantly, Channel \ours\ trained on available tasks and instructions still achieves lower performance than Channel \ours\ without templates/instructions ($46.9$ from Table~\ref{tab:ablate_inst} vs. $49.1$ from Table~\ref{tab:main-result}). This is likely because the model with instructions was trained with less meta-training tasks, which was unavoidable since instructions are only available on 32 out of 61 meta-training tasks. This supports our earlier choice of not using human-written templates/instructions, since writing templates and instructions for every task requires extensive effort.

It is worth noting that, it is nonetheless difficult to make direct comparisons with \citet{wei2022finetuned} and \citet{sanh2022multitask} because there are many moving components: size of LMs, types of LMs (e.g., causal LM vs. masked LM), splits between meta-training and target tasks, and more.

\section{Conclusion}\label{sec:concl}In this paper, we introduced \ours, a new few-shot learning method where an LM is meta-trained to learn to in-context learn, i.e. condition on training examples to recover the task and make predictions.
%
We experiment with a large, diverse collection of tasks, consisting of 142 unique tasks in total and 52 unique target tasks, using seven different settings.
\ours\ outperforms a range of strong baselines including in-context learning without meta-training and multi-task learning followed by zero-shot transfer, and outperforms or matches 8x bigger models.
We identify ingredients for success of \ours\ such as the number and diversity of meta-training tasks.
We also demonstrate that, while \ours\ is better than recent work using natural instructions, they are complementary and the best performance is achieved by integrating \ours\ with instructions.

\paragraph{Limitation \& Future work}
Our work is limited in multiple dimensions. First, in-context learning approaches in general requires much longer context at both meta-training and inference due to feeding the concatenation of the training data, thus being less efficient compared to baselines that do not use in-context learning.
Second, our work experiment with a casual language model with modest size (GPT-2 Large, 770M parameters). Future work may investigate extending our approach to a masked language model and a larger model.
Third, our experiments focus on classification and multi-choice tasks where a set of candidate options is given. Future work may study applying our approach for a wider range of tasks including free-form generation.
Other avenues for future work include further improving \ours\ to outperform supervised models with meta-training, identification of which meta-training tasks are helpful on target tasks, and how to better combine human-written instructions and \ours. 

\section*{Acknowledgements}
We thank Ari Holtzman and Victoria Lin for comments and discussions, and Tim Dettmers for help with experiments.
This research was supported by NSF IIS-2044660, ONR N00014-18-1-2826, 
an Allen Distinguished Investigator Award, and a Sloan Fellowship. 

\bibliography{acl,datasets}

\begin{thebibliography}{128}
\expandafter\ifx\csname natexlab\endcsname\relax\def\natexlab#1{#1}\fi

\bibitem[{Aghajanyan et~al.(2021)Aghajanyan, Gupta, Shrivastava, Chen,
  Zettlemoyer, and Gupta}]{aghajanyan2021muppet}
Armen Aghajanyan, Anchit Gupta, Akshat Shrivastava, Xilun Chen, Luke
  Zettlemoyer, and Sonal Gupta. 2021.
\newblock Muppet: Massive multi-task representations with pre-finetuning.
\newblock \emph{arXiv preprint arXiv:2101.11038}.

\bibitem[{Almeida et~al.(2011)Almeida, Hidalgo, and Yamakami}]{sms_spam}
Tiago~A. Almeida, Jos\'{e} Mar\'{\i}a~G. Hidalgo, and Akebo Yamakami. 2011.
\newblock Contributions to the study of sms spam filtering: New collection and
  results.
\newblock In \emph{Proceedings of the 11th ACM Symposium on Document
  Engineering}.

\bibitem[{Bar-Haim et~al.(2006)Bar-Haim, Dagan, Dolan, Ferro, Giampiccolo,
  Magnini, and Szpektor}]{bar2006second}
Roy Bar-Haim, Ido Dagan, Bill Dolan, Lisa Ferro, Danilo Giampiccolo, Bernardo
  Magnini, and Idan Szpektor. 2006.
\newblock The second pascal recognising textual entailment challenge.
\newblock In \emph{Proceedings of the second PASCAL challenges workshop on
  recognising textual entailment}.

\bibitem[{Barbieri et~al.(2020)Barbieri, Camacho-Collados, Espinosa~Anke, and
  Neves}]{barbieri-etal-2020-tweeteval}
Francesco Barbieri, Jose Camacho-Collados, Luis Espinosa~Anke, and Leonardo
  Neves. 2020.
\newblock {T}weet{E}val: Unified benchmark and comparative evaluation for tweet
  classification.
\newblock In \emph{Findings of the Association for Computational Linguistics:
  EMNLP 2020}.

\bibitem[{Bentivogli et~al.(2009)Bentivogli, Clark, Dagan, and
  Giampiccolo}]{bentivogli2009fifth}
Luisa Bentivogli, Peter Clark, Ido Dagan, and Danilo Giampiccolo. 2009.
\newblock The fifth pascal recognizing textual entailment challenge.
\newblock In \emph{TAC}.

\bibitem[{Berant et~al.(2013)Berant, Chou, Frostig, and
  Liang}]{berant-etal-2013-semantic}
Jonathan Berant, Andrew Chou, Roy Frostig, and Percy Liang. 2013.
\newblock Semantic parsing on {F}reebase from question-answer pairs.
\newblock In \emph{EMNLP}.

\bibitem[{Bhagavatula et~al.(2020)Bhagavatula, Bras, Malaviya, Sakaguchi,
  Holtzman, Rashkin, Downey, tau Yih, and Choi}]{bhagavatula2020abductive}
Chandra Bhagavatula, Ronan~Le Bras, Chaitanya Malaviya, Keisuke Sakaguchi, Ari
  Holtzman, Hannah Rashkin, Doug Downey, Wen tau Yih, and Yejin Choi. 2020.
\newblock Abductive commonsense reasoning.
\newblock In \emph{ICLR}.

\bibitem[{Bisk et~al.(2020)Bisk, Zellers, Bras, Gao, and Choi}]{bisk2019piqa}
Yonatan Bisk, Rowan Zellers, Ronan~Le Bras, Jianfeng Gao, and Yejin Choi. 2020.
\newblock Piqa: Reasoning about physical commonsense in natural language.
\newblock In \emph{AAAI}.

\bibitem[{Boratko et~al.(2020)Boratko, Li, O{'}Gorman, Das, Le, and
  McCallum}]{boratko-etal-2020-protoqa}
Michael Boratko, Xiang Li, Tim O{'}Gorman, Rajarshi Das, Dan Le, and Andrew
  McCallum. 2020.
\newblock {P}roto{QA}: A question answering dataset for prototypical
  common-sense reasoning.
\newblock In \emph{EMNLP}.

\bibitem[{Brown et~al.(2020)Brown, Mann, Ryder, Subbiah, Kaplan, Dhariwal,
  Neelakantan, Shyam, Sastry, Askell, Agarwal, Herbert-Voss, Krueger, Henighan,
  Child, Ramesh, Ziegler, Wu, Winter, Hesse, Chen, Sigler, Litwin, Gray, Chess,
  Clark, Berner, McCandlish, Radford, Sutskever, and
  Amodei}]{brown2020language}
Tom Brown, Benjamin Mann, Nick Ryder, Melanie Subbiah, Jared~D Kaplan, Prafulla
  Dhariwal, Arvind Neelakantan, Pranav Shyam, Girish Sastry, Amanda Askell,
  Sandhini Agarwal, Ariel Herbert-Voss, Gretchen Krueger, Tom Henighan, Rewon
  Child, Aditya Ramesh, Daniel Ziegler, Jeffrey Wu, Clemens Winter, Chris
  Hesse, Mark Chen, Eric Sigler, Mateusz Litwin, Scott Gray, Benjamin Chess,
  Jack Clark, Christopher Berner, Sam McCandlish, Alec Radford, Ilya Sutskever,
  and Dario Amodei. 2020.
\newblock Language models are few-shot learners.
\newblock In \emph{NeurIPS}.

\bibitem[{Carlini et~al.(2021)Carlini, Tramer, Wallace, Jagielski,
  Herbert-Voss, Lee, Roberts, Brown, Song, Erlingsson
  et~al.}]{carlini2021extracting}
Nicholas Carlini, Florian Tramer, Eric Wallace, Matthew Jagielski, Ariel
  Herbert-Voss, Katherine Lee, Adam Roberts, Tom Brown, Dawn Song, Ulfar
  Erlingsson, et~al. 2021.
\newblock Extracting training data from large language models.
\newblock In \emph{30th USENIX Security Symposium (USENIX Security 21)}.

\bibitem[{Chatterjee et~al.(2019)Chatterjee, Narahari, Joshi, and
  Agrawal}]{chatterjee-etal-2019-semeval}
Ankush Chatterjee, Kedhar~Nath Narahari, Meghana Joshi, and Puneet Agrawal.
  2019.
\newblock {S}em{E}val-2019 task 3: {E}mo{C}ontext contextual emotion detection
  in text.
\newblock In \emph{Proceedings of the 13th International Workshop on Semantic
  Evaluation}.

\bibitem[{Chen et~al.(2019)Chen, D{'}Arcy, Liu, Fernandez, and
  Downey}]{chen-etal-2019-codah}
Michael Chen, Mike D{'}Arcy, Alisa Liu, Jared Fernandez, and Doug Downey. 2019.
\newblock {CODAH}: An adversarially-authored question answering dataset for
  common sense.
\newblock In \emph{Proceedings of the 3rd Workshop on Evaluating Vector Space
  Representations for {NLP}}.

\bibitem[{Chen et~al.(2020)Chen, Wang, Chen, Zhang, Wang, Li, Zhou, and
  Wang}]{Chen2020TabFact}
Wenhu Chen, Hongmin Wang, Jianshu Chen, Yunkai Zhang, Hong Wang, Shiyang Li,
  Xiyou Zhou, and William~Yang Wang. 2020.
\newblock Tabfact: A large-scale dataset for table-based fact verification.
\newblock In \emph{ICLR}.

\bibitem[{Chen et~al.(2022)Chen, Zhong, Zha, Karypis, and He}]{chen2022meta}
Yanda Chen, Ruiqi Zhong, Sheng Zha, George Karypis, and He~He. 2022.
\newblock Meta-learning via language model in-context tuning.
\newblock In \emph{ACL}.

\bibitem[{Clark et~al.(2019)Clark, Lee, Chang, Kwiatkowski, Collins, and
  Toutanova}]{clark-etal-2019-boolq}
Christopher Clark, Kenton Lee, Ming-Wei Chang, Tom Kwiatkowski, Michael
  Collins, and Kristina Toutanova. 2019.
\newblock {B}ool{Q}: Exploring the surprising difficulty of natural yes/no
  questions.
\newblock In \emph{NAACL-HLT}.

\bibitem[{Clark et~al.(2018)Clark, Cowhey, Etzioni, Khot, Sabharwal, Schoenick,
  and Tafjord}]{Clark2018ThinkYH}
Peter Clark, Isaac Cowhey, Oren Etzioni, Tushar Khot, Ashish Sabharwal, Carissa
  Schoenick, and Oyvind Tafjord. 2018.
\newblock Think you have solved question answering? try arc, the ai2 reasoning
  challenge.
\newblock \emph{arXiv preprint arXiv:1803.05457}.

\bibitem[{Dagan et~al.(2005)Dagan, Glickman, and Magnini}]{dagan2005pascal}
Ido Dagan, Oren Glickman, and Bernardo Magnini. 2005.
\newblock The pascal recognising textual entailment challenge.
\newblock In \emph{Machine Learning Challenges Workshop}.

\bibitem[{Dasigi et~al.(2019)Dasigi, Liu, Marasovi{\'c}, Smith, and
  Gardner}]{dasigi-etal-2019-quoref}
Pradeep Dasigi, Nelson~F. Liu, Ana Marasovi{\'c}, Noah~A. Smith, and Matt
  Gardner. 2019.
\newblock {Q}uoref: A reading comprehension dataset with questions requiring
  coreferential reasoning.
\newblock In \emph{EMNLP}.

\bibitem[{Davidson et~al.(2017)Davidson, Warmsley, Macy, and
  Weber}]{hateoffensive}
Thomas Davidson, Dana Warmsley, Michael Macy, and Ingmar Weber. 2017.
\newblock Automated hate speech detection and the problem of offensive
  language.
\newblock In \emph{Proceedings of the 11th International AAAI Conference on Web
  and Social Media}.

\bibitem[{de~Gibert et~al.(2018)de~Gibert, Perez, Garc{\'\i}a-Pablos, and
  Cuadros}]{gibert2018hate}
Ona de~Gibert, Naiara Perez, Aitor Garc{\'\i}a-Pablos, and Montse Cuadros.
  2018.
\newblock {Hate Speech Dataset from a White Supremacy Forum}.
\newblock In \emph{Proceedings of the 2nd Workshop on Abusive Language Online
  ({ALW}2)}.

\bibitem[{de~Marneffe et~al.(2019)de~Marneffe, Simons, and
  Tonhauser}]{Marneffe_Simons_Tonhauser_2019}
Marie-Catherine de~Marneffe, Mandy Simons, and Judith Tonhauser. 2019.
\newblock The commitmentbank: Investigating projection in naturally occurring
  discourse.
\newblock \emph{Proceedings of Sinn und Bedeutung}.

\bibitem[{Dettmers et~al.(2022)Dettmers, Lewis, Shleifer, and
  Zettlemoyer}]{dettmers20228}
Tim Dettmers, Mike Lewis, Sam Shleifer, and Luke Zettlemoyer. 2022.
\newblock 8-bit optimizers via block-wise quantization.
\newblock In \emph{ICLR}.

\bibitem[{Diggelmann et~al.(2020)Diggelmann, Boyd-Graber, Bulian, Ciaramita,
  and Leippold}]{Diggelmann2020CLIMATEFEVERAD}
T.~Diggelmann, Jordan~L. Boyd-Graber, Jannis Bulian, Massimiliano Ciaramita,
  and Markus Leippold. 2020.
\newblock Climate-fever: A dataset for verification of real-world climate
  claims.
\newblock \emph{ArXiv}.

\bibitem[{Dolan and Brockett(2005)}]{dolan-brockett-2005-automatically}
William~B. Dolan and Chris Brockett. 2005.
\newblock Automatically constructing a corpus of sentential paraphrases.
\newblock In \emph{Proceedings of the Third International Workshop on
  Paraphrasing ({IWP}2005)}.

\bibitem[{Dua et~al.(2019)Dua, Wang, Dasigi, Stanovsky, Singh, and
  Gardner}]{dua-etal-2019-drop}
Dheeru Dua, Yizhong Wang, Pradeep Dasigi, Gabriel Stanovsky, Sameer Singh, and
  Matt Gardner. 2019.
\newblock {DROP}: {A} reading comprehension benchmark requiring discrete
  reasoning over paragraphs.
\newblock In \emph{NAACL}.

\bibitem[{Dunn et~al.(2017)Dunn, Sagun, Higgins, G{\"u}ney, Cirik, and
  Cho}]{Dunn2017SearchQAAN}
Matthew Dunn, Levent Sagun, Mike Higgins, V.~U. G{\"u}ney, Volkan Cirik, and
  Kyunghyun Cho. 2017.
\newblock Searchqa: A new q\&a dataset augmented with context from a search
  engine.
\newblock \emph{arXiv preprint arXiv:1704.05179}.

\bibitem[{Elsahar et~al.(2018)Elsahar, Vougiouklis, Remaci, Gravier, Hare,
  Laforest, and Simperl}]{elsahar-etal-2018-rex}
Hady Elsahar, Pavlos Vougiouklis, Arslen Remaci, Christophe Gravier, Jonathon
  Hare, Frederique Laforest, and Elena Simperl. 2018.
\newblock T-{RE}x: A large scale alignment of natural language with knowledge
  base triples.
\newblock In \emph{LREC}.

\bibitem[{Evgeniou and Pontil(2004)}]{evgeniou2004regularized}
Theodoros Evgeniou and Massimiliano Pontil. 2004.
\newblock Regularized multi--task learning.
\newblock In \emph{Proceedings of the tenth ACM SIGKDD international conference
  on Knowledge discovery and data mining}.

\bibitem[{Faruqui and Das(2018)}]{faruqui-das-2018-identifying}
Manaal Faruqui and Dipanjan Das. 2018.
\newblock Identifying well-formed natural language questions.
\newblock In \emph{EMNLP}.

\bibitem[{Finn et~al.(2017)Finn, Abbeel, and Levine}]{finn2017model}
Chelsea Finn, Pieter Abbeel, and Sergey Levine. 2017.
\newblock Model-agnostic meta-learning for fast adaptation of deep networks.
\newblock In \emph{ICML}.

\bibitem[{Giampiccolo et~al.(2007)Giampiccolo, Magnini, Dagan, and
  Dolan}]{giampiccolo2007third}
Danilo Giampiccolo, Bernardo Magnini, Ido Dagan, and Bill Dolan. 2007.
\newblock The third pascal recognizing textual entailment challenge.
\newblock In \emph{Proceedings of the ACL-PASCAL workshop on textual entailment
  and paraphrasing}.

\bibitem[{Gordon et~al.(2012)Gordon, Kozareva, and
  Roemmele}]{gordon-etal-2012-semeval}
Andrew Gordon, Zornitsa Kozareva, and Melissa Roemmele. 2012.
\newblock {S}em{E}val-2012 task 7: Choice of plausible alternatives: An
  evaluation of commonsense causal reasoning.
\newblock In \emph{The First Joint Conference on Lexical and Computational
  Semantics ({S}em{E}val)}.

\bibitem[{Gurulingappa et~al.(2012)Gurulingappa, Rajput, Roberts, Fluck,
  Hofmann-Apitius, and Toldo}]{GURULINGAPPA2012885}
Harsha Gurulingappa, Abdul~Mateen Rajput, Angus Roberts, Juliane Fluck, Martin
  Hofmann-Apitius, and Luca Toldo. 2012.
\newblock Development of a benchmark corpus to support the automatic extraction
  of drug-related adverse effects from medical case reports.
\newblock \emph{Journal of Biomedical Informatics}.

\bibitem[{He et~al.(2015)He, Lewis, and Zettlemoyer}]{he-etal-2015-question}
Luheng He, Mike Lewis, and Luke Zettlemoyer. 2015.
\newblock Question-answer driven semantic role labeling: Using natural language
  to annotate natural language.
\newblock In \emph{Proceedings of the 2015 Conference on Empirical Methods in
  Natural Language Processing}.

\bibitem[{Hoffart et~al.(2011)Hoffart, Yosef, Bordino, F{\"u}rstenau, Pinkal,
  Spaniol, Taneva, Thater, and Weikum}]{hoffart-etal-2011-robust}
Johannes Hoffart, Mohamed~Amir Yosef, Ilaria Bordino, Hagen F{\"u}rstenau,
  Manfred Pinkal, Marc Spaniol, Bilyana Taneva, Stefan Thater, and Gerhard
  Weikum. 2011.
\newblock Robust disambiguation of named entities in text.
\newblock In \emph{EMNLP}.

\bibitem[{Holtzman et~al.(2021)Holtzman, West, Schwartz, Choi, and
  Zettlemoyer}]{holtzman2021surface}
Ari Holtzman, Peter West, Vered Schwartz, Yejin Choi, and Luke Zettlemoyer.
  2021.
\newblock Surface form competition: Why the highest probability answer isn't
  always right.
\newblock In \emph{EMNLP}.

\bibitem[{Hovy et~al.(2001)Hovy, Gerber, Hermjakob, Lin, and
  Ravichandran}]{hovy-etal-2001-toward}
Eduard Hovy, Laurie Gerber, Ulf Hermjakob, Chin-Yew Lin, and Deepak
  Ravichandran. 2001.
\newblock Toward semantics-based answer pinpointing.
\newblock In \emph{Proceedings of the First International Conference on Human
  Language Technology Research}.

\bibitem[{Huang et~al.(2019)Huang, Le~Bras, Bhagavatula, and
  Choi}]{huang-etal-2019-cosmos}
Lifu Huang, Ronan Le~Bras, Chandra Bhagavatula, and Yejin Choi. 2019.
\newblock Cosmos {QA}: Machine reading comprehension with contextual
  commonsense reasoning.
\newblock In \emph{EMNLP}.

\bibitem[{Jiang et~al.(2019)Jiang, Wu, and Jiang}]{jiang-etal-2019-freebaseqa}
Kelvin Jiang, Dekun Wu, and Hui Jiang. 2019.
\newblock {F}reebase{QA}: A new factoid {QA} data set matching trivia-style
  question-answer pairs with {F}reebase.
\newblock In \emph{NAACL-HLT}.

\bibitem[{Khashabi et~al.(2018)Khashabi, Chaturvedi, Roth, Upadhyay, and
  Roth}]{khashabi-etal-2018-looking}
Daniel Khashabi, Snigdha Chaturvedi, Michael Roth, Shyam Upadhyay, and Dan
  Roth. 2018.
\newblock Looking beyond the surface: A challenge set for reading comprehension
  over multiple sentences.
\newblock In \emph{NAACL-HLT}.

\bibitem[{Khashabi et~al.(2020)Khashabi, Min, Khot, Sabharwal, Tafjord, Clark,
  and Hajishirzi}]{khashabi2020unifiedqa}
Daniel Khashabi, Sewon Min, Tushar Khot, Ashish Sabharwal, Oyvind Tafjord,
  Peter Clark, and Hannaneh Hajishirzi. 2020.
\newblock {U}nified{QA}: Crossing format boundaries with a single qa system.
\newblock In \emph{Findings of EMNLP}.

\bibitem[{Khot et~al.(2019)Khot, Clark, Guerquin, Jansen, and
  Sabharwal}]{khot2019qasc}
Tushar Khot, Peter Clark, Michal Guerquin, Peter Jansen, and Ashish Sabharwal.
  2019.
\newblock {QASC}: A dataset for question answering via sentence composition.
\newblock In \emph{AAAI}.

\bibitem[{Khot et~al.(2020)Khot, Clark, Guerquin, Jansen, and
  Sabharwal}]{Khot_Clark_Guerquin_Jansen_Sabharwal_2020}
Tushar Khot, Peter Clark, Michal Guerquin, Peter Jansen, and Ashish Sabharwal.
  2020.
\newblock Qasc: A dataset for question answering via sentence composition.
\newblock In \emph{AAAI}.

\bibitem[{Khot et~al.(2018)Khot, Sabharwal, and Clark}]{scitail}
Tushar Khot, Ashish Sabharwal, and Peter Clark. 2018.
\newblock Scitail: A textual entailment dataset from science question
  answering.
\newblock In \emph{AAAI}.

\bibitem[{Kingma and Ba(2015)}]{kingma2015adam}
Diederik~P Kingma and Jimmy Ba. 2015.
\newblock Adam: A method for stochastic optimization.
\newblock In \emph{ICLR}.

\bibitem[{Kocisk{\'y} et~al.(2018)Kocisk{\'y}, Schwarz, Blunsom, Dyer, Hermann,
  Melis, and Grefenstette}]{kocisky-etal-2018-narrativeqa}
Tom{\'a}s Kocisk{\'y}, Jonathan Schwarz, Phil Blunsom, Chris Dyer, Karl~Moritz
  Hermann, G{\'a}bor Melis, and Edward Grefenstette. 2018.
\newblock The narrativeqa reading comprehension challenge.
\newblock \emph{TACL}.

\bibitem[{Kotonya and Toni(2020)}]{kotonya-toni-2020-explainable-automated}
Neema Kotonya and Francesca Toni. 2020.
\newblock Explainable automated fact-checking for public health claims.
\newblock In \emph{EMNLP}.

\bibitem[{Kwiatkowski et~al.(2019)Kwiatkowski, Palomaki, Redfield, Collins,
  Parikh, Alberti, Epstein, Polosukhin, Devlin, Lee, Toutanova, Jones, Kelcey,
  Chang, Dai, Uszkoreit, Le, and Petrov}]{kwiatkowski-etal-2019-natural}
Tom Kwiatkowski, Jennimaria Palomaki, Olivia Redfield, Michael Collins,
  Ankur~P. Parikh, Chris Alberti, Danielle Epstein, Illia Polosukhin, Jacob
  Devlin, Kenton Lee, Kristina Toutanova, Llion Jones, Matthew Kelcey, Ming-Wei
  Chang, Andrew~M. Dai, Jakob Uszkoreit, Quoc Le, and Slav Petrov. 2019.
\newblock Natural questions: A benchmark for question answering research.
\newblock \emph{TACL}.

\bibitem[{Lai et~al.(2017)Lai, Xie, Liu, Yang, and Hovy}]{lai-etal-2017-race}
Guokun Lai, Qizhe Xie, Hanxiao Liu, Yiming Yang, and Eduard~H. Hovy. 2017.
\newblock {RACE}: {L}arge-scale reading comprehension dataset from
  examinations.
\newblock In \emph{EMNLP}.

\bibitem[{Lehmann et~al.(2015)Lehmann, Isele, Jakob, Jentzsch, Kontokostas,
  Mendes, Hellmann, Morsey, van Kleef, Auer, and Bizer}]{Lehmann2015DBpediaA}
Jens Lehmann, Robert Isele, Max Jakob, Anja Jentzsch, D.~Kontokostas, Pablo~N.
  Mendes, Sebastian Hellmann, M.~Morsey, Patrick van Kleef, S.~Auer, and
  C.~Bizer. 2015.
\newblock Dbpedia - a large-scale, multilingual knowledge base extracted from
  wikipedia.
\newblock \emph{Semantic Web}.

\bibitem[{Levesque et~al.(2012)Levesque, Davis, and
  Morgenstern}]{levesque2012winograd}
Hector~J. Levesque, Ernest Davis, and Leora Morgenstern. 2012.
\newblock The winograd schema challenge.
\newblock In \emph{Proceedings of the Thirteenth International Conference on
  Principles of Knowledge Representation and Reasoning}.

\bibitem[{Levy et~al.(2017)Levy, Seo, Choi, and
  Zettlemoyer}]{levy-etal-2017-zero}
Omer Levy, Minjoon Seo, Eunsol Choi, and Luke Zettlemoyer. 2017.
\newblock Zero-shot relation extraction via reading comprehension.
\newblock In \emph{{C}o{NLL}}.

\bibitem[{Li and Roth(2002)}]{li-roth-2002-learning}
Xin Li and Dan Roth. 2002.
\newblock Learning question classifiers.
\newblock In \emph{COLING}.

\bibitem[{Lin et~al.(2020)Lin, Lee, Khanna, and Ren}]{lin-etal-2020-birds}
Bill~Yuchen Lin, Seyeon Lee, Rahul Khanna, and Xiang Ren. 2020.
\newblock {B}irds have four legs?! {N}umer{S}ense: {P}robing {N}umerical
  {C}ommonsense {K}nowledge of {P}re-{T}rained {L}anguage {M}odels.
\newblock In \emph{EMNLP}.

\bibitem[{Lin et~al.(2019)Lin, Tafjord, Clark, and
  Gardner}]{lin-etal-2019-reasoning}
Kevin Lin, Oyvind Tafjord, Peter Clark, and Matt Gardner. 2019.
\newblock Reasoning over paragraph effects in situations.
\newblock In \emph{Proceedings of the 2nd Workshop on Machine Reading for
  Question Answering}.

\bibitem[{Louis et~al.(2020)Louis, Roth, and Radlinski}]{louis-etal-2020-id}
Annie Louis, Dan Roth, and Filip Radlinski. 2020.
\newblock {``}{I}{'}d rather just go to bed{''}: Understanding indirect
  answers.
\newblock In \emph{EMNLP}.

\bibitem[{Lu et~al.(2021)Lu, Bartolo, Moore, Riedel, and
  Stenetorp}]{lu2021fantastically}
Yao Lu, Max Bartolo, Alastair Moore, Sebastian Riedel, and Pontus Stenetorp.
  2021.
\newblock Fantastically ordered prompts and where to find them: Overcoming
  few-shot prompt order sensitivity.
\newblock \emph{arXiv preprint arXiv:2104.08786}.

\bibitem[{Maas et~al.(2011)Maas, Daly, Pham, Huang, Ng, and
  Potts}]{maas-etal-2011-learning}
Andrew~L. Maas, Raymond~E. Daly, Peter~T. Pham, Dan Huang, Andrew~Y. Ng, and
  Christopher Potts. 2011.
\newblock Learning word vectors for sentiment analysis.
\newblock In \emph{Proceedings of the 49th Annual Meeting of the Association
  for Computational Linguistics: Human Language Technologies}.

\bibitem[{Malo et~al.(2014)Malo, Sinha, Korhonen, Wallenius, and
  Takala}]{financial-phrasebank}
Pekka Malo, Ankur Sinha, Pekka Korhonen, Jyrki Wallenius, and Pyry Takala.
  2014.
\newblock Good debt or bad debt: Detecting semantic orientations in economic
  texts.
\newblock \emph{J. Assoc. Inf. Sci. Technol.}

\bibitem[{Marelli et~al.(2014)Marelli, Menini, Baroni, Bentivogli, Bernardi,
  and Zamparelli}]{marelli-etal-2014-sick}
Marco Marelli, Stefano Menini, Marco Baroni, Luisa Bentivogli, Raffaella
  Bernardi, and Roberto Zamparelli. 2014.
\newblock A {SICK} cure for the evaluation of compositional distributional
  semantic models.
\newblock In \emph{LREC}.

\bibitem[{Mathew et~al.(2020)Mathew, Saha, Yimam, Biemann, Goyal, and
  Mukherjee}]{mathew2020hatexplain}
Binny Mathew, Punyajoy Saha, Seid~Muhie Yimam, Chris Biemann, Pawan Goyal, and
  Animesh Mukherjee. 2020.
\newblock Hatexplain: A benchmark dataset for explainable hate speech
  detection.
\newblock \emph{arXiv preprint arXiv:2012.10289}.

\bibitem[{McAuley and Leskovec(2013)}]{McAuley2013HiddenFA}
Julian McAuley and J.~Leskovec. 2013.
\newblock Hidden factors and hidden topics: understanding rating dimensions
  with review text.
\newblock \emph{Proceedings of the 7th ACM conference on Recommender systems}.

\bibitem[{McCreery et~al.(2020)McCreery, Katariya, Kannan, Chablani, and
  Amatriain}]{medical-qqp}
Clara~H. McCreery, Namit Katariya, Anitha Kannan, Manish Chablani, and Xavier
  Amatriain. 2020.
\newblock Effective transfer learning for identifying similar questions:
  Matching user questions to covid-19 faqs.
\newblock In \emph{Proceedings of the 26th ACM SIGKDD International Conference
  on Knowledge Discovery \& Data Mining}.

\bibitem[{Micikevicius et~al.(2017)Micikevicius, Narang, Alben, Diamos, Elsen,
  Garcia, Ginsburg, Houston, Kuchaiev, Venkatesh
  et~al.}]{micikevicius2017mixed}
Paulius Micikevicius, Sharan Narang, Jonah Alben, Gregory Diamos, Erich Elsen,
  David Garcia, Boris Ginsburg, Michael Houston, Oleksii Kuchaiev, Ganesh
  Venkatesh, et~al. 2017.
\newblock Mixed precision training.
\newblock In \emph{ICLR}.

\bibitem[{Mihaylov et~al.(2018)Mihaylov, Clark, Khot, and
  Sabharwal}]{mihaylov-etal-2018-suit}
Todor Mihaylov, Peter Clark, Tushar Khot, and Ashish Sabharwal. 2018.
\newblock Can a suit of armor conduct electricity? a new dataset for open book
  question answering.
\newblock In \emph{EMNLP}.

\bibitem[{Min et~al.(2022)Min, Lewis, Hajishirzi, and
  Zettlemoyer}]{min2022noisy}
Sewon Min, Mike Lewis, Hannaneh Hajishirzi, and Luke Zettlemoyer. 2022.
\newblock Noisy channel language model prompting for few-shot text
  classification.
\newblock In \emph{ACL}.

\bibitem[{Mishra et~al.(2021)Mishra, Khashabi, Baral, Choi, and
  Hajishirzi}]{mishra2021reframing}
Swaroop Mishra, Daniel Khashabi, Chitta Baral, Yejin Choi, and Hannaneh
  Hajishirzi. 2021.
\newblock Reframing instructional prompts to gptk's language.
\newblock \emph{arXiv preprint arXiv:2109.07830}.

\bibitem[{Mishra et~al.(2022)Mishra, Khashabi, Baral, and
  Hajishirzi}]{mishra2022cross}
Swaroop Mishra, Daniel Khashabi, Chitta Baral, and Hannaneh Hajishirzi. 2022.
\newblock Cross-task generalization via natural language crowdsourcing
  instructions.
\newblock In \emph{ACL}.

\bibitem[{Mollas et~al.(2020)Mollas, Chrysopoulou, Karlos, and
  Tsoumakas}]{Mollas2020ETHOSAO}
Ioannis Mollas, Zoe Chrysopoulou, Stamatis Karlos, and Grigorios Tsoumakas.
  2020.
\newblock Ethos: an online hate speech detection dataset.
\newblock \emph{arXiv preprint arXiv:2006.08328}.

\bibitem[{Nangia et~al.(2020)Nangia, Vania, Bhalerao, and
  Bowman}]{nangia-etal-2020-crows}
Nikita Nangia, Clara Vania, Rasika Bhalerao, and Samuel~R. Bowman. 2020.
\newblock {C}row{S}-pairs: A challenge dataset for measuring social biases in
  masked language models.
\newblock In \emph{EMNLP}.

\bibitem[{Napoles et~al.(2012)Napoles, Gormley, and
  Van~Durme}]{napoles-etal-2012-annotated}
Courtney Napoles, Matthew Gormley, and Benjamin Van~Durme. 2012.
\newblock Annotated {G}igaword.
\newblock In \emph{Proceedings of the Joint Workshop on Automatic Knowledge
  Base Construction and Web-scale Knowledge Extraction ({AKBC}-{WEKEX})}.

\bibitem[{Narayan et~al.(2018)Narayan, Cohen, and
  Lapata}]{narayan-etal-2018-dont}
Shashi Narayan, Shay~B. Cohen, and Mirella Lapata. 2018.
\newblock Don{'}t give me the details, just the summary! topic-aware
  convolutional neural networks for extreme summarization.
\newblock In \emph{EMNLP}.

\bibitem[{Nie et~al.(2020)Nie, Williams, Dinan, Bansal, Weston, and
  Kiela}]{nie-etal-2020-adversarial}
Yixin Nie, Adina Williams, Emily Dinan, Mohit Bansal, Jason Weston, and Douwe
  Kiela. 2020.
\newblock Adversarial {NLI}: A new benchmark for natural language
  understanding.
\newblock In \emph{ACL}.

\bibitem[{Pang and Lee(2005)}]{pang-lee-2005-seeing}
Bo~Pang and Lillian Lee. 2005.
\newblock Seeing stars: Exploiting class relationships for sentiment
  categorization with respect to rating scales.
\newblock In \emph{ACL}.

\bibitem[{Pappas et~al.(2020)Pappas, Stavropoulos, Androutsopoulos, and
  McDonald}]{pappas-etal-2020-biomrc}
Dimitris Pappas, Petros Stavropoulos, Ion Androutsopoulos, and Ryan McDonald.
  2020.
\newblock {B}io{MRC}: A dataset for biomedical machine reading comprehension.
\newblock In \emph{Proceedings of the 19th SIGBioMed Workshop on Biomedical
  Language Processing}.

\bibitem[{Paszke et~al.(2019)Paszke, Gross, Massa, Lerer, Bradbury, Chanan,
  Killeen, Lin, Gimelshein, Antiga et~al.}]{paszke2019pytorch}
Adam Paszke, Sam Gross, Francisco Massa, Adam Lerer, James Bradbury, Gregory
  Chanan, Trevor Killeen, Zeming Lin, Natalia Gimelshein, Luca Antiga, et~al.
  2019.
\newblock Pytorch: An imperative style, high-performance deep learning library.
\newblock In \emph{NeurIPS}.

\bibitem[{Perez et~al.(2021)Perez, Kiela, and Cho}]{perez2021true}
Ethan Perez, Douwe Kiela, and Kyunghyun Cho. 2021.
\newblock True few-shot learning with language models.
\newblock In \emph{NeurIPS}.

\bibitem[{Petroni et~al.(2020)Petroni, Lewis, Piktus, Rockt{\"a}schel, Wu,
  Miller, and Riedel}]{petroni2020how}
Fabio Petroni, Patrick Lewis, Aleksandra Piktus, Tim Rockt{\"a}schel, Yuxiang
  Wu, Alexander~H. Miller, and Sebastian Riedel. 2020.
\newblock How context affects language models' factual predictions.
\newblock In \emph{Automated Knowledge Base Construction}.

\bibitem[{Petroni et~al.(2019)Petroni, Rockt{\"a}schel, Riedel, Lewis, Bakhtin,
  Wu, and Miller}]{petroni-etal-2019-language}
Fabio Petroni, Tim Rockt{\"a}schel, Sebastian Riedel, Patrick Lewis, Anton
  Bakhtin, Yuxiang Wu, and Alexander Miller. 2019.
\newblock Language models as knowledge bases?
\newblock In \emph{EMNLP}.

\bibitem[{Pilehvar and
  Camacho-Collados(2019)}]{pilehvar-camacho-collados-2019-wic}
Mohammad~Taher Pilehvar and Jose Camacho-Collados. 2019.
\newblock {W}i{C}: the word-in-context dataset for evaluating context-sensitive
  meaning representations.
\newblock In \emph{NAACL-HLT}.

\bibitem[{Radford et~al.(2019)Radford, Wu, Child, Luan, Amodei, and
  Sutskever}]{radford2019language}
Alec Radford, Jeffrey Wu, Rewon Child, David Luan, Dario Amodei, and Ilya
  Sutskever. 2019.
\newblock Language models are unsupervised multitask learners.
\newblock \emph{OpenAI blog}.

\bibitem[{Rajpurkar et~al.(2018)Rajpurkar, Jia, and
  Liang}]{rajpurkar-etal-2018-know}
Pranav Rajpurkar, Robin Jia, and Percy Liang. 2018.
\newblock Know what you don't know: Unanswerable questions for squad.
\newblock In \emph{ACL}.

\bibitem[{Rajpurkar et~al.(2016)Rajpurkar, Zhang, Lopyrev, and
  Liang}]{rajpurkar-etal-2016-squad}
Pranav Rajpurkar, Jian Zhang, Konstantin Lopyrev, and Percy Liang. 2016.
\newblock {SQuAD}: 100,000+ questions for machine comprehension of text.
\newblock In \emph{{EMNLP}}.

\bibitem[{Richardson et~al.(2013)Richardson, Burges, and
  Renshaw}]{richardson-etal-2013-mctest}
Matthew Richardson, Christopher J.~C. Burges, and Erin Renshaw. 2013.
\newblock Mctest: A challenge dataset for the open-domain machine comprehension
  of text.
\newblock In \emph{EMNLP}.

\bibitem[{Rogers et~al.(2020)Rogers, Kovaleva, Downey, and
  Rumshisky}]{Rogers_Kovaleva_Downey_Rumshisky_2020}
Anna Rogers, Olga Kovaleva, Matthew Downey, and Anna Rumshisky. 2020.
\newblock Getting closer to ai complete question answering: A set of
  prerequisite real tasks.
\newblock In \emph{AAAI}.

\bibitem[{Ruder(2017)}]{ruder2017overview}
Sebastian Ruder. 2017.
\newblock An overview of multi-task learning in deep neural networks.
\newblock \emph{arXiv preprint arXiv:1706.05098}.

\bibitem[{Sakaguchi et~al.(2020{\natexlab{a}})Sakaguchi, Bras, Bhagavatula, and
  Choi}]{sakaguchi2019winogrande}
Keisuke Sakaguchi, Ronan~Le Bras, Chandra Bhagavatula, and Yejin Choi.
  2020{\natexlab{a}}.
\newblock {WINOGRANDE:} an adversarial winograd schema challenge at scale.
\newblock In \emph{AAAI}.

\bibitem[{Sakaguchi et~al.(2020{\natexlab{b}})Sakaguchi, Le~Bras, Bhagavatula,
  and Choi}]{Sakaguchi_Le_Bras_Bhagavatula_Choi_2020}
Keisuke Sakaguchi, Ronan Le~Bras, Chandra Bhagavatula, and Yejin Choi.
  2020{\natexlab{b}}.
\newblock Winogrande: An adversarial winograd schema challenge at scale.
\newblock In \emph{AAAI}.

\bibitem[{Sanh et~al.(2022)Sanh, Webson, Raffel, Bach, Sutawika, Alyafeai,
  Chaffin, Stiegler, Scao, Raja, Dey, Bari, Xu, Thakker, Sharma, Szczechla,
  Kim, Chhablani, Nayak, Datta, Chang, Jiang, Wang, Manica, Shen, Yong, Pandey,
  Bawden, Wang, Neeraj, Rozen, Sharma, Santilli, Fevry, Fries, Teehan,
  Biderman, Gao, Bers, Wolf, and Rush}]{sanh2022multitask}
Victor Sanh, Albert Webson, Colin Raffel, Stephen~H. Bach, Lintang Sutawika,
  Zaid Alyafeai, Antoine Chaffin, Arnaud Stiegler, Teven~Le Scao, Arun Raja,
  Manan Dey, M~Saiful Bari, Canwen Xu, Urmish Thakker, Shanya Sharma, Eliza
  Szczechla, Taewoon Kim, Gunjan Chhablani, Nihal Nayak, Debajyoti Datta,
  Jonathan Chang, Mike Tian-Jian Jiang, Han Wang, Matteo Manica, Sheng Shen,
  Zheng~Xin Yong, Harshit Pandey, Rachel Bawden, Thomas Wang, Trishala Neeraj,
  Jos Rozen, Abheesht Sharma, Andrea Santilli, Thibault Fevry, Jason~Alan
  Fries, Ryan Teehan, Stella Biderman, Leo Gao, Tali Bers, Thomas Wolf, and
  Alexander~M. Rush. 2022.
\newblock Multitask prompted training enables zero-shot task generalization.
\newblock In \emph{ICLR}.

\bibitem[{Sap et~al.(2019{\natexlab{a}})Sap, Rashkin, Chen, Le~Bras, and
  Choi}]{sap-etal-2019-social}
Maarten Sap, Hannah Rashkin, Derek Chen, Ronan Le~Bras, and Yejin Choi.
  2019{\natexlab{a}}.
\newblock Social {IQ}a: Commonsense reasoning about social interactions.
\newblock In \emph{EMNLP}.

\bibitem[{Sap et~al.(2019{\natexlab{b}})Sap, Rashkin, Chen, Le~Bras, and
  Choi}]{sap2019socialiqa}
Maarten Sap, Hannah Rashkin, Derek Chen, Ronan Le~Bras, and Yejin Choi.
  2019{\natexlab{b}}.
\newblock Social iqa: Commonsense reasoning about social interactions.
\newblock In \emph{EMNLP-IJCNLP}.

\bibitem[{Saravia et~al.(2018)Saravia, Liu, Huang, Wu, and
  Chen}]{saravia-etal-2018-carer}
Elvis Saravia, Hsien-Chi~Toby Liu, Yen-Hao Huang, Junlin Wu, and Yi-Shin Chen.
  2018.
\newblock {CARER}: Contextualized affect representations for emotion
  recognition.
\newblock In \emph{EMNLP}.

\bibitem[{Sheng and Uthus(2020)}]{sheng-uthus-2020-investigating}
Emily Sheng and David Uthus. 2020.
\newblock Investigating societal biases in a poetry composition system.
\newblock In \emph{Proceedings of the Second Workshop on Gender Bias in Natural
  Language Processing}.

\bibitem[{Sileo et~al.(2019)Sileo, Van De~Cruys, Pradel, and
  Muller}]{sileo-etal-2019-mining}
Damien Sileo, Tim Van De~Cruys, Camille Pradel, and Philippe Muller. 2019.
\newblock Mining discourse markers for unsupervised sentence representation
  learning.
\newblock In \emph{NAACL-HLT}.

\bibitem[{Socher et~al.(2013)Socher, Perelygin, Wu, Chuang, Manning, Ng, and
  Potts}]{socher-etal-2013-recursive}
Richard Socher, Alex Perelygin, Jean Wu, Jason Chuang, Christopher~D. Manning,
  Andrew Ng, and Christopher Potts. 2013.
\newblock Recursive deep models for semantic compositionality over a sentiment
  treebank.
\newblock In \emph{EMNLP}.

\bibitem[{Sun et~al.(2019)Sun, Yu, Chen, Yu, Choi, and
  Cardie}]{sun-etal-2019-dream}
Kai Sun, Dian Yu, Jianshu Chen, Dong Yu, Yejin Choi, and Claire Cardie. 2019.
\newblock {DREAM}: A challenge data set and models for dialogue-based reading
  comprehension.
\newblock \emph{TACL}.

\bibitem[{Tafjord et~al.(2019{\natexlab{a}})Tafjord, Clark, Gardner, Yih, and
  Sabharwal}]{Tafjord_Clark_Gardner_Yih_Sabharwal_2019}
Oyvind Tafjord, Peter Clark, Matt Gardner, Wen-tau Yih, and Ashish Sabharwal.
  2019{\natexlab{a}}.
\newblock Quarel: A dataset and models for answering questions about
  qualitative relationships.
\newblock In \emph{AAAI}.

\bibitem[{Tafjord et~al.(2019{\natexlab{b}})Tafjord, Gardner, Lin, and
  Clark}]{tafjord-etal-2019-quartz}
Oyvind Tafjord, Matt Gardner, Kevin Lin, and Peter Clark. 2019{\natexlab{b}}.
\newblock {Q}ua{RT}z: An open-domain dataset of qualitative relationship
  questions.
\newblock In \emph{EMNLP}.

\bibitem[{Talmor et~al.(2019)Talmor, Herzig, Lourie, and
  Berant}]{talmor-etal-2019-commonsenseqa}
Alon Talmor, Jonathan Herzig, Nicholas Lourie, and Jonathan Berant. 2019.
\newblock Commonsenseqa: A question answering challenge targeting commonsense
  knowledge.
\newblock In \emph{NAACL-HLT}.

\bibitem[{Tandon et~al.(2019)Tandon, Dalvi, Sakaguchi, Clark, and
  Bosselut}]{tandon-etal-2019-wiqa}
Niket Tandon, Bhavana Dalvi, Keisuke Sakaguchi, Peter Clark, and Antoine
  Bosselut. 2019.
\newblock {WIQA}: A dataset for {``}what if...{''} reasoning over procedural
  text.
\newblock In \emph{EMNLP}.

\bibitem[{Thorne et~al.(2018)Thorne, Vlachos, Christodoulopoulos, and
  Mittal}]{thorne-etal-2018-fever}
James Thorne, Andreas Vlachos, Christos Christodoulopoulos, and Arpit Mittal.
  2018.
\newblock {FEVER}: a large-scale dataset for fact extraction and
  {VER}ification.
\newblock In \emph{NAACL-HLT}.

\bibitem[{Trischler et~al.(2017)Trischler, Wang, Yuan, Harris, Sordoni,
  Bachman, and Suleman}]{trischler-etal-2017-newsqa}
Adam Trischler, Tong Wang, Xingdi Yuan, Justin Harris, Alessandro Sordoni,
  Philip Bachman, and Kaheer Suleman. 2017.
\newblock Newsqa: A machine comprehension dataset.
\newblock In \emph{Rep4NLP@ACL}.

\bibitem[{Vajjala and
  Lu{\v{c}}i{\'c}(2018)}]{vajjala-lucic-2018-onestopenglish}
Sowmya Vajjala and Ivana Lu{\v{c}}i{\'c}. 2018.
\newblock {O}ne{S}top{E}nglish corpus: A new corpus for automatic readability
  assessment and text simplification.
\newblock In \emph{Proceedings of the Thirteenth Workshop on Innovative Use of
  {NLP} for Building Educational Applications}.

\bibitem[{Vilalta and Drissi(2002)}]{vilalta2002perspective}
Ricardo Vilalta and Youssef Drissi. 2002.
\newblock A perspective view and survey of meta-learning.
\newblock \emph{Artificial intelligence review}.

\bibitem[{Wang et~al.(2018)Wang, Singh, Michael, Hill, Levy, and
  Bowman}]{wang2018glue}
Alex Wang, Amanpreet Singh, Julian Michael, Felix Hill, Omer Levy, and Samuel~R
  Bowman. 2018.
\newblock Glue: A multi-task benchmark and analysis platform for natural
  language understanding.
\newblock In \emph{{B}lackbox{NLP} Workshop: Analyzing and Interpreting Neural
  Networks for {NLP}}.

\bibitem[{Wang and Komatsuzaki(2021)}]{wang2021gpt}
Ben Wang and Aran Komatsuzaki. 2021.
\newblock {GPT-J-6B: A 6 Billion Parameter Autoregressive Language Model}.
\newblock \url{https://github.com/kingoflolz/mesh-transformer-jax}.

\bibitem[{Wang(2017)}]{wang-2017-liar}
William~Yang Wang. 2017.
\newblock {``}liar, liar pants on fire{''}: A new benchmark dataset for fake
  news detection.
\newblock In \emph{ACL}.

\bibitem[{Warstadt et~al.(2020)Warstadt, Parrish, Liu, Mohananey, Peng, Wang,
  and Bowman}]{warstadt2019blimp}
Alex Warstadt, Alicia Parrish, Haokun Liu, Anhad Mohananey, Wei Peng, Sheng-Fu
  Wang, and Samuel~R. Bowman. 2020.
\newblock Blimp: The benchmark of linguistic minimal pairs for english.
\newblock \emph{TACL}.

\bibitem[{Warstadt et~al.(2019)Warstadt, Singh, and
  Bowman}]{warstadt-etal-2019-neural}
Alex Warstadt, Amanpreet Singh, and Samuel~R. Bowman. 2019.
\newblock Neural network acceptability judgments.
\newblock \emph{TACL}.

\bibitem[{Wei et~al.(2022)Wei, Bosma, Zhao, Guu, Yu, Lester, Du, Dai, and
  Le}]{wei2022finetuned}
Jason Wei, Maarten Bosma, Vincent~Y Zhao, Kelvin Guu, Adams~Wei Yu, Brian
  Lester, Nan Du, Andrew~M Dai, and Quoc~V Le. 2022.
\newblock Finetuned language models are zero-shot learners.
\newblock In \emph{ICLR}.

\bibitem[{Welbl et~al.(2017)Welbl, Liu, and
  Gardner}]{welbl-etal-2017-crowdsourcing}
Johannes Welbl, Nelson~F. Liu, and Matt Gardner. 2017.
\newblock Crowdsourcing multiple choice science questions.
\newblock In \emph{Proceedings of the 3rd Workshop on Noisy User-generated
  Text}.

\bibitem[{Williams et~al.(2018)Williams, Nangia, and
  Bowman}]{williams-etal-2018-broad}
Adina Williams, Nikita Nangia, and Samuel Bowman. 2018.
\newblock A broad-coverage challenge corpus for sentence understanding through
  inference.
\newblock In \emph{NAACL-HLT}.

\bibitem[{Wolf et~al.(2020)Wolf, Debut, Sanh, Chaumond, Delangue, Moi, Cistac,
  Rault, Louf, Funtowicz, Davison, Shleifer, von Platen, Ma, Jernite, Plu, Xu,
  Scao, Gugger, Drame, Lhoest, and Rush}]{wolf-etal-2020-transformers}
Thomas Wolf, Lysandre Debut, Victor Sanh, Julien Chaumond, Clement Delangue,
  Anthony Moi, Pierric Cistac, Tim Rault, Rémi Louf, Morgan Funtowicz, Joe
  Davison, Sam Shleifer, Patrick von Platen, Clara Ma, Yacine Jernite, Julien
  Plu, Canwen Xu, Teven~Le Scao, Sylvain Gugger, Mariama Drame, Quentin Lhoest,
  and Alexander~M. Rush. 2020.
\newblock Transformers: State-of-the-art natural language processing.
\newblock In \emph{EMNLP: System Demonstrations}.

\bibitem[{Xiong et~al.(2019)Xiong, Wu, Wang, Kulkarni, Yu, Chang, Guo, and
  Wang}]{xiong-etal-2019-tweetqa}
Wenhan Xiong, Jiawei Wu, Hong Wang, Vivek Kulkarni, Mo~Yu, Shiyu Chang,
  Xiaoxiao Guo, and William~Yang Wang. 2019.
\newblock {TWEETQA}: A social media focused question answering dataset.
\newblock In \emph{ACL}.

\bibitem[{Yang et~al.(2015)Yang, Yih, and Meek}]{yang-etal-2015-wikiqa}
Yi~Yang, Wen-tau Yih, and Christopher Meek. 2015.
\newblock {W}iki{QA}: A challenge dataset for open-domain question answering.
\newblock In \emph{EMNLP}.

\bibitem[{Yang et~al.(2018)Yang, Qi, Zhang, Bengio, Cohen, Salakhutdinov, and
  Manning}]{yang-etal-2018-hotpotqa}
Zhilin Yang, Peng Qi, Saizheng Zhang, Yoshua Bengio, William Cohen, Ruslan
  Salakhutdinov, and Christopher~D. Manning. 2018.
\newblock {H}otpot{QA}: A dataset for diverse, explainable multi-hop question
  answering.
\newblock In \emph{EMNLP}.

\bibitem[{Ye et~al.(2021)Ye, Lin, and Ren}]{ye2021crossfit}
Qinyuan Ye, Bill~Yuchen Lin, and Xiang Ren. 2021.
\newblock Crossfit: A few-shot learning challenge for cross-task generalization
  in nlp.
\newblock In \emph{EMNLP}.

\bibitem[{Yu et~al.(2018)Yu, Zhang, Yang, Yasunaga, Wang, Li, Ma, Li, Yao,
  Roman, Zhang, and Radev}]{yu-etal-2018-spider}
Tao Yu, Rui Zhang, Kai Yang, Michihiro Yasunaga, Dongxu Wang, Zifan Li, James
  Ma, Irene Li, Qingning Yao, Shanelle Roman, Zilin Zhang, and Dragomir Radev.
  2018.
\newblock {S}pider: A large-scale human-labeled dataset for complex and
  cross-domain semantic parsing and text-to-{SQL} task.
\newblock In \emph{EMNLP}.

\bibitem[{Zellers et~al.(2018)Zellers, Bisk, Schwartz, and
  Choi}]{zellers-etal-2018-swag}
Rowan Zellers, Yonatan Bisk, Roy Schwartz, and Yejin Choi. 2018.
\newblock {SWAG}: A large-scale adversarial dataset for grounded commonsense
  inference.
\newblock In \emph{EMNLP}.

\bibitem[{Zellers et~al.(2019)Zellers, Holtzman, Bisk, Farhadi, and
  Choi}]{zellers-etal-2019-hellaswag}
Rowan Zellers, Ari Holtzman, Yonatan Bisk, Ali Farhadi, and Yejin Choi. 2019.
\newblock {H}ella{S}wag: Can a machine really finish your sentence?
\newblock In \emph{ACL}.

\bibitem[{Zhang et~al.(2018)Zhang, Liu, Liu, Gao, Duh, and
  Durme}]{Zhang2018ReCoRDBT}
Sheng Zhang, X.~Liu, J.~Liu, Jianfeng Gao, Kevin Duh, and Benjamin~Van Durme.
  2018.
\newblock Record: Bridging the gap between human and machine commonsense
  reading comprehension.
\newblock \emph{arXiv preprint arXiv:1810.12885}.

\bibitem[{Zhang et~al.(2015)Zhang, Zhao, and LeCun}]{zhang2015character}
Xiang Zhang, Junbo Zhao, and Yann LeCun. 2015.
\newblock Character-level convolutional networks for text classification.
\newblock In \emph{NeurIPS}.

\bibitem[{Zhang et~al.(2019)Zhang, Baldridge, and He}]{zhang-etal-2019-paws}
Yuan Zhang, Jason Baldridge, and Luheng He. 2019.
\newblock {PAWS}: Paraphrase adversaries from word scrambling.
\newblock In \emph{NAACL-HLT}.

\bibitem[{Zhao et~al.(2021)Zhao, Wallace, Feng, Klein, and
  Singh}]{zhao2021calibrate}
Tony~Z Zhao, Eric Wallace, Shi Feng, Dan Klein, and Sameer Singh. 2021.
\newblock Calibrate before use: Improving few-shot performance of language
  models.
\newblock In \emph{ICML}.

\bibitem[{Zhong et~al.(2021)Zhong, Lee, Zhang, and Klein}]{zhong2021adapting}
Ruiqi Zhong, Kristy Lee, Zheng Zhang, and Dan Klein. 2021.
\newblock Adapting language models for zero-shot learning by meta-tuning on
  dataset and prompt collections.
\newblock In \emph{Findings of EMNLP}.

\bibitem[{Zhong et~al.(2017)Zhong, Xiong, and Socher}]{zhongSeq2SQL2017}
Victor Zhong, Caiming Xiong, and Richard Socher. 2017.
\newblock Seq2sql: Generating structured queries from natural language using
  reinforcement learning.
\newblock \emph{arXiv preprint arXiv:1709.00103}.

\bibitem[{Zhou et~al.(2019)Zhou, Khashabi, Ning, and
  Roth}]{zhou-etal-2019-going}
Ben Zhou, Daniel Khashabi, Qiang Ning, and Dan Roth. 2019.
\newblock {``}going on a vacation{''} takes longer than {``}going for a
  walk{''}: A study of temporal commonsense understanding.
\newblock In \emph{EMNLP}.

\end{thebibliography}
\bibliographystyle{acl_natbib}

\clearpage
\appendix
\section{Dataset List}
\label{app:dataset}

Table~\ref{tab:full-datasets} and Table~\ref{tab:full-datasets-citations} report a list of datasets used in the settings detailed in Section~\ref{subsec:dataset}.
The first 10 rows are for settings described in Section~\ref{subsec:dataset}; the next two rows are for settings used for ablations on the diversity of meta-training tasks (Table~\ref{tab:ablate_diversity} of Section~\ref{subsec:ablations}); the last two rows are for settings used for ablations on using natural instructions (Table~\ref{tab:ablate_inst} of Section~\ref{subsec:ablations}).
\textbf{Bold} datasets are target datasets with no overlap in domain with meta-training tasks.
All datasets are taken from \textsc{CrossFit}~\citep{ye2021crossfit} (except we exclude datasets that are unavailable from their repository\footnote{
\href{https://github.com/INK-USC/CrossFit}{\nolinkurl{github.com/INK-USC/CrossFit}}} or the scope is notably different from other tasks, e.g., solving math problems or breaking down compositional questions) and \textsc{UnifiedQA}~\citep{khashabi2020unifiedqa}.

\paragraph{How meta-training/target splits are determined}
The \main\ setting is created based on the training data size as described in Section~\ref{subsec:dataset}. Settings involving Classification, NLI and Paraphrase are taken from \textsc{CrossFit}. Settings involving QA are created by combining QA datasets from \textsc{CrossFit} and datasets from \textsc{UnifiedQA}.

\vspace{.2em}
Statistics are reported in Table~\ref{tab:data-summary} and Table~\ref{tab:data-statistics}.
The number of tasks is the largest among recent related work: we have 142 unique tasks, while \citet{khashabi2020unifiedqa}, \citet{zhong2021adapting}, \citet{mishra2022cross},  \citet{wei2022finetuned} and \citet{sanh2022multitask} use 32, 62, 61, 42 and 62 tasks, respectively.
References for all datasets are provided in Table~\ref{tab:full-datasets-citations}.
Data and splits are available at \code.

\section{Implementation Details}
\label{app:impl-details}
\paragraph{Preprocessing details}
For all models with meta-training and the raw GPT-J, we separate the input and the output with one newline (\texttt{$\backslash$n}), and separate between examples with three newlines. For the raw GPT-2, we use spaces instead of newlines. This choice was made in order to report the best baseline performance we were able to achieve: when raw LMs are used, GPT-2 is significantly better with spaces than with newlines, and GPT-J is significantly better with newlines than with spaces.\footnote{For example, in the \main\ setting, the raw GPT-2 is about $4$\% better with spaces then with newlines, and the raw GPT-J is about $5$\% better with spaces and then with newlines (all with the channel in-context learning method).}
We note that \ours\ is less sensitive to these formatting differences, having less than 2\% differences between using spaces and using newlines.

When the concatenation of $k$ examples is too long, we truncate each example to have at most $256$ tokens, and truncate the earlier tokens of the concatenation so that the LM sees the recent tokens. Additionally, for extractive question answering datasets as meta-training tasks, the input passage is truncated with a guarantee that the groundtruth answer is included in the input passage. We do not do this truncation for target datasets.

\begin{table}[t]
    \centering \footnotesize
    \setlength{\tabcolsep}{0.4em}
    \begin{tabular}{lrrrr}
        \toprule
            \multirow{2}{*}{Setting} & 
            \multicolumn{2}{c}{Input} & \multicolumn{2}{c}{Output} \\
            \cmidrule(lr){2-3} \cmidrule(lr){4-5}
            & Mean & Median & Mean & Median \\
        \midrule
            \multicolumn{5}{c}{\em Meta-training tasks} \\
            HR                  & 81.7 & 73 & 2.8 & 2 \\
            Classification      & 45.8 & 41 & 1.1 & 1 \\
            Non-Classification  & 77.7 & 69 & 4.2 & 3 \\
            QA                  & 142.6 & 137 & 2.7 & 2 \\
            Non-QA              & 68.7 & 56 & 2.3 & 2 \\
            Non-NLI             & 44.3 & 39 & 1.1 & 1 \\
            Non-Paraphrase      & 45.0 & 39 & 1.1 & 1 \\
        \midrule
            \multicolumn{5}{c}{\em Target tasks} \\
            LR                  & 29.7 & 25 & 1.9 & 1 \\
            Classification      & 44.9 & 38 & 1.0 & 1 \\
            QA                  & 74.4 & 69 & 4.6 & 4 \\
            NLI                 & 45.4 & 41 & 1.0 & 1 \\
            Paraphrase          & 42.2 & 41 & 1.0 & 1 \\
        \bottomrule
    \end{tabular}
    \caption{Length statistics of tasks used in different settings, before any truncation.
    We compute the mean and the median of each task, and report the macro-average over all tasks for each setting.
    }\label{tab:data-statistics}
\end{table}

\begin{table*}[!ht]
    \centering \footnotesize
    \setlength{\tabcolsep}{0.6em}
    \begin{tabular}{
        l @{\hspace{3em}} ccccc
        }
        \toprule
            Method & \main & 
            \makecell[c]{\{Class,non-Class\} \\ $\rightarrow$Class} & \makecell[c]{\{QA,non-QA\} \\ $\rightarrow$QA} &
            \makecell[c]{non-NLI \\ $\rightarrow$NLI} &
            \makecell[c]{non-Para \\ $\rightarrow$Para} \\
        \midrule
            \multicolumn{6}{c}{\em All tasks} \\
            0-shot & 31.5 & 31.5 & 45.6 & 25.7 & 30.0 \\
            PMI 0-shot & 36.9 & 30.2 & 44.3 & 30.2 & 37.6 \\
            Channel 0-shot & 39.7 & 41.5 & 42.1 & 36.2 & 45.0 \\
            In-context & 43.8/39.1 & 43.6/34.3 & \textbf{50.8}/48.3 & 35.0/27.6 & 41.3/33.2 \\
            PMI In-context & 43.0/37.4 & 44.8/36.6 & 48.8/46.9 & 31.5/26.0 & 38.4/33.6 \\
            Channel In-context & \textbf{48.6}/44.4 & \textbf{51.5}/47.0 & 47.0/45.2 & \textbf{47.2}/41.7 & \textbf{51.0}/47.5 \\
        \midrule
            \multicolumn{6}{c}{\em Target tasks in unseen domains} \\
            0-shot & 31.2 & 31.2 & 47.5 & 33.5 & 34.1 \\
            PMI 0-shot & 25.2 & 25.2 & 43.8 & 36.1 & 34.4 \\
            Channel 0-shot & 37.2 & 37.2 & 46.9 & \textbf{53.4} & \textbf{54.7} \\
            In-context & 33.1/25.4 & 33.1/25.4 & \textbf{57.4}/53.1 & 46.7/36.1 & 34.1/34.1 \\
            PMI In-context & 35.4/28.2 & 35.4/28.2 & 54.5/50.9 & 33.9/33.9 & 32.5/32.4 \\
            Channel In-context & \textbf{42.8}/38.4 & \textbf{42.8}/38.4 & 55.7/54.4 & 51.1/40.4 & 52.0/46.5 \\
        \bottomrule
    \end{tabular}\vspace{-.3em}
    \caption{Performance of raw LM baselines using \textbf{GPT-J} (6B).
    Two numbers indicate the average and the worst-case accuracy over different seeds used for $k$ target training examples.
    `Class' indicate `Classification'.
    }\label{tab:full-result}
\end{table*}

\paragraph{Comparison with baselines in training and inference cost} Although being trained for the same global steps (30,000 steps), it takes 3 hours to train Multi-task 0-shot baselines (in contrast to 4.5 hours for \ours), likely because the sequence length is 4x shorter. At inference, Multi-task 0-shot baselines are roughly 4x more efficient, also because the sequence length is 4x shorter.\footnote{Let $L$ be the sequence length, the memory requirement for attention layers and feed-forward layers are $O(L^2)$ and $O(L)$, respectively. In practice, feed-forward layers are responsible for most memory usage when the size of the transformers is large, thus empirical memory usage tends to be linear to $L$.} We did not control for the training time and the inference time for comparison since both models are efficient enough.

\begin{table*}[!ht]
    \centering \footnotesize
    \setlength{\tabcolsep}{.6em}
    \begin{tabular}{l@{\hspace{2em}} cccc c cccc}
        \toprule
            & \multicolumn{4}{c}{\em All tasks} && \multicolumn{4}{c}{\em Target tasks in unseen domains} \\
        \cmidrule{2-5} \cmidrule{7-10}
            & S & M & L & XL && S & M & L & XL \\
        \midrule
            Channel In-context  & 41.5/37.4 & 42.2/37.7 & 43.1/38.5 & 43.5/39.9 && 40.9/35.9 & 38.8/34.7 & 39.6/33.6 & 40.0/37.2\\
            MT 0-shot           & 35.4 & 36.4 & 35.6 & - && 34.9 & 32.2 & 35.4 & -\\
            Channel MT 0-shot   & 40.4 & 37.9 & 38.8 & - && 33.8 & 35.9 & 36.3 & -\\
            \ours & 39.7/36.2   & 40.3/36.4 & 43.3/41.7 & - &&  36.9/32.6 & 38.1/35.0 & 35.3/32.7 & - \\
            Channel \ours       & 46.2/43.1 & 44.3/41.5 & 49.1/46.8 & - && 46.9/42.6 & 43.1/39.8 & 47.7/44.7 & - \\
        \bottomrule
    \end{tabular}
    \caption{Ablation on the size of the LM on the \main\ setting.
    We use small, medium, large, and XL variants of GPT-2.
    We were unable to meta-train the XL variant due to memory limit.
    }\label{tab:ablate_size}
\end{table*}

\paragraph{Ablations in using instructions}
When we choose one instruction per task at meta-training tasks, we choose one by (1) first excluding the instruction if its name contains \texttt{no\_option}, (2) then taking the instruction which name contains \texttt{multiple\_choice}, \texttt{most\_correct} or \texttt{most\_suitable} if there are any, and (3) if not, then randomly sampling one. We choose one instruction per target task at test time using the same process. This is different \citet{sanh2022multitask} where the median of the performance over all instructions is reported. We think our choice better reflects the real use-case scenario---choosing one instruction that looks the most reasonable to human.
    
\section{Additional Results \& Analyses}
\label{app:results}

\subsection{GPT-J results}\label{app:gpt-j-result}
Table~\ref{tab:full-result} reports the full results of raw LM baselines based on GPT-J,
consisting of 6B parameters. See Section~\ref{subsec:main-results} for discussion.

\subsection{Varying LM sizes}\label{app:abl_lm_size}
We vary the size of the GPT-2 models---small, medium, large, and XL---with 124M, 355M, 774M, and 1.5B parameters, respectively. Results are reported in Table~\ref{tab:ablate_size}. We find that (1) increasing the model size generally helps, (2) for all model sizes, Channel \ours\ significantly outperforms baselines, and (3) \ours\ enables a much smaller model to outperform a bigger model, e.g., Channel MetaICL based on GPT-2 Small outperforms the GPT-2 XL baseline that is 12x bigger (46.2 vs. 43.5).

\begin{table*}[!th]
    \centering \footnotesize
    \begin{tabular}{l @{\hspace{0.5em}} p{0.85\textwidth}}
        \toprule
            \em Single task \\
            Helpful: & tweet\_eval-offensive, glue-sst2, glue-mnli, wino\_grande, kilt\_hotpotqa \\
            Unhelpful: & race-middle, cosmos\_qa, dbpedia\_14,  gigaword, wikisql \\
        \midrule
            \em Task pair \\
            Helpful: &
            (yelp\_review\_full, glue-mnli), (yelp\_review\_full, wino\_grande),
            (hateexplain, glue-sst2),
            (hateexplain, glue-mnli),
            (hateexplain, glue-qqp),
            \\
            Unhelpful: &
            (paws, dbpedia\_14),
            (paws, art),
            (paws, cosmos\_qa),
            (cosmos\_qa, dbpedia\_14),
            (quail, art)
            \\
        \midrule
            \em Task triple \\
            Helpful &
            (yelp\_review\_full, glue-qqp, glue-mnli),
            (yelp\_review\_full, glue-sst2, glue-mnli),
            (yelp\_review\_full, hateexplain, glue-mnli),
            (yelp\_review\_full, hateexplain, qqp),
            (yelp\_review\_full, hate\_speech\_offensive, glue-mnli),
            \\
            Unhelpful &
            (paws, dbpedia\_14, art),
            (paws, dbpedia\_14, cosmos\_qa),
            (paws, cosmos\_qa, art),
            (dbpedia\_14, cosmos\_qa, art),
            (quail, paws, dbpedia\_14)
            \\
        \bottomrule
    \end{tabular}\vspace{-.3em}
    \caption{Analysis of which meta-training tasks give good performance in Channel \ours.
    We report five most helpful and the most unhelpful tasks (or task sets), respectively.}\label{tab:analysis_task}
\end{table*}

\subsection{Which meta-training tasks are more helpful?}\label{app:which-tasks-helpful}

Based on large variance across different choices of meta-training (Figure~\ref{fig:ablate_size} of Section~\ref{subsec:ablations}), we think certain tasks are more helpful for meta-training than other tasks. In this context, we create $50$ sets of seven meta-training tasks using $50$ different random seeds. We then measure the correlation between tasks/task pairs/task triples and average performance of Channel \ours\ when the task is included in the meta-training tasks.

Table~\ref{tab:analysis_task} reports the result.
We first find that high quality datasets with diverse domain like GLUE family~\citep{wang2018glue} are often helpful.
We also find that datasets that are collected adversarially (e.g. \texttt{paws}, \texttt{art}) or are notably dissimilar from all other tasks (e.g. \texttt{wikisql} that requires semantic parsing) are often unhelpful.
Nonetheless, we were not able to find good explanations for other cases, e.g., many sentiment analysis datasets being particularly helpful even though only 3 out of 26 target datasets are sentiment analysis, and \texttt{dbpedia\_14}/\texttt{cosmos\_qa}/\texttt{race-middle} being unhelpful.
Moreover, we think which tasks are helpful largely depends on the choice of target tasks, and we should not make early conclusions that certain tasks are helpful/unhelpful in all cases.
We think future work should investigate these impacts in a more systematic way.


\begin{table}[t]
    \centering \footnotesize
    \begin{tabular}{
        l @{\hspace{.7em}} lc @{\hspace{1em}} c}
        \toprule
            \multirow{2}{*}{Method} & \multirow{2}{*}{Train labels} & \multicolumn{2}{c}{Test labels} \\
            \cmidrule{3-4}
            && Original & Replaced \\
        \midrule
            \multicolumn{4}{c}{\em All target tasks} \\
            \multicolumn{2}{l}{Random} & 36.0 & 36.0 \\
            \multicolumn{2}{l}{0-shot} & 34.2 & 23.8/16.8 \\
            \multicolumn{2}{l}{Channel 0-shot} & 37.3 & 31.4/22.9 \\
            \multicolumn{2}{l}{In-context} & 37.4/33.9 & 30.5/25.0 \\
            \multicolumn{2}{l}{Channel In-context} & 46.3/40.3 & 37.7/31.3 \\
        \cmidrule{1-4}
            MT 0-shot & Original & 37.3 & 25.5/16.4 \\
            Channel MT 0-shot & Original & 40.9 & 28.6/19.9 \\
            \ours & Original & 43.4/39.9 & 30.1/24.0 \\
            Channel \ours & Original & \textbf{50.7}/48.0 & 36.5/28.9 \\
        \cmidrule{1-4}
            MT 0-shot & Replaced & 24.4 & 23.1/15.5 \\
            Channel MT 0-shot & Replaced & 36.7 & 34.1/28.4 \\
            \ours & Replaced & 40.7/36.0 & \textbf{43.5}/35.2 \\
            Channel \ours & Replaced & 47.1/42.7 & 39.5/33.7 \\
        \bottomrule
    \end{tabular}\vspace{-.3em}
    \caption{Ablation where label words are replaced with {\em random English word} in the class$\rightarrow$class setting.
    {\em Original} and {\em Replaced} indicate original label words and labels that are replaced to random English words, respectively.
    When tested on {\em Replaced}, five random seeds used to sample English words.
    }\label{tab:ablate_labels}
\end{table}

\subsection{Does \ours\ generalize when semantic hints from label words are removed?}
Our experiments use label words taken from the original dataset, which often contain {\em semantic hints}---hints on what each label is supposed to mean (\texttt{entailment} and \texttt{not\_entailment} for the NLI task, and \texttt{positive} and \texttt{negative} for the sentiment analysis task).
If the model is truly learning the task in-context, it should generalize when label words are replaced with random English words, e.g., \texttt{entailment} and \texttt{not\_entailment} are replaced with \texttt{apple} and \texttt{orange}, thus not giving any hints about the task.
In this context, we run experiments where each label word is replaced with a random word sampled from 61,569 common English words.\footnote{
\href{https://pypi.org/project/english-words}{\nolinkurl{pypi.org/project/english-words}}.}
We use five seeds for sampling random words, and report the average and the worst-case performance. 

Results in Table~\ref{tab:ablate_labels} show that raw LMs (the first block of the table) and models trained on the original data (the second block) achieve near random guessing performance. This indicates that having semantic hints from label words is a necessary condition for all models to perform the task. 

Next, we meta-train the MT 0-shot baseline and \ours\ where, for each iteration of meta-training, we similarly map label words with random words. The mapping from the label set to sampled English words is independent for each iteration, so that the model never sees the same mapping during meta-training and hence does not overfit to a specific mapping.
Results are reported in the third block of Table~\ref{tab:ablate_labels}.
MT 0-shot baselines are still not better than random guessing, which is expected as they have no way to grasp the meaning of each label.
On the other hand, \ours\ benefits from training on the replaced data, improving performance from 30.1\% to 43.5\% while retaining most performance on the original data ($43.4\%\rightarrow40.7\%$).

Still, overall performance is relatively poor.
We think
future work should investigate the model that can in-context learn {\em any} task.

\begin{table*}[!t]
    \centering \scriptsize
    \begin{tabular}{p{\textwidth}}
        \toprule
            Setting: \main\ Meta-train \\
            piqa, hate\_speech\_offensive, google\_wellformed\_query, social\_i\_qa, circa, quoref, glue-sst2, scitail, emo, cosmos\_qa, freebase\_qa, ag\_news, art, paws,
            kilt\_ay2, glue-qnli, quail, ade\_corpus\_v2-classification, sciq, hatexplain, emotion, glue-qqp, kilt\_fever, kilt\_nq, dbpedia\_14, kilt\_zsre, hellaswag, squad-with\_context,
            hotpot\_qa, glue-mnli, ropes, squad-no\_context, kilt\_hotpotqa, discovery, superglue-record, race-middle, race-high, lama-trex, swag, gigaword, amazon\_polarity,
            biomrc, tab\_fact, tweet\_eval-emoji, tweet\_eval-offensive, tweet\_eval-sentiment, tweet\_qa, imdb, lama-conceptnet, liar, anli, wiki\_qa, kilt\_trex, wikisql, wino\_grande,
            wiqa, search\_qa, xsum, yahoo\_answers\_topics, yelp\_polarity, yelp\_review\_full \\
        \midrule
            Setting: \main\ Target \\
            quarel, \textbf{financial\_phrasebank}, openbookqa, codah, qasc, glue-mrpc, dream, sick, commonsense\_qa, \textbf{medical\_questions\_pairs}, quartz-with\_knowledge,
            \textbf{poem\_sentiment}, quartz-no\_knowledge, glue-wnli, \textbf{climate\_fever}, ethos-national\_origin, ethos-race, ethos-religion, ai2\_arc, hate\_speech18,
            glue-rte, superglue-cb, superglue-copa, tweet\_eval-hate, tweet\_eval-stance\_atheism, tweet\_eval-stance\_feminist \\
        \midrule
            Setting: Classification Meta-train \\
            Meta-Train: superglue-rte, tweet\_eval-sentiment, discovery, glue-rte, superglue-wsc, glue-mrpc, tweet\_eval-stance\_hillary, tweet\_eval-offensive,
            emotion, hatexplain, glue-cola, sick, paws, ethos-sexual\_orientation, glue-qqp, tweet\_eval-emotion, sms\_spam, health\_fact, glue-mnli, imdb, ethos-disability,
            glue-wnli, scitail, trec-finegrained, yahoo\_answers\_topics, liar, glue-sst2, tweet\_eval-stance\_abortion, circa, tweet\_eval-stance\_climate, glue-qnli, tweet\_eval-emoji,
            ethos-directed\_vs\_generalized, ade\_corpus\_v2-classification, hate\_speech\_offensive, superglue-wic, google\_wellformed\_query, tweet\_eval-irony,
            ethos-gender, onestop\_english, trec, rotten\_tomatoes, kilt\_fever \\
        \midrule
            Setting: Non-Classification Meta-train \\
            ade\_corpus\_v2-dosage, art, biomrc, blimp-anaphor\_number\_agreement, blimp-ellipsis\_n\_bar\_2, blimp-sentential\_negation\_npi\_licensor\_present,
            blimp-sentential\_negation\_npi\_scope, commonsense\_qa, crows\_pairs, dream, freebase\_qa, gigaword, hellaswag, hotpot\_qa, kilt\_ay2, kilt\_hotpotqa, kilt\_trex,
            kilt\_zsre, lama-conceptnet, lama-google\_re, lama-squad, numer\_sense, openbookqa, piqa, proto\_qa, qa\_srl, quarel, quartz-no\_knowledge, race-high, ropes, sciq,
            social\_i\_qa, spider, superglue-multirc, wikisql, xsum, yelp\_review\_full
            \\
        \midrule
            Setting: Classification Target \\
            tweet\_eval-stance\_feminist, ethos-national\_origin, tweet\_eval-hate, ag\_news, amazon\_polarity, hate\_speech18, \textbf{poem\_sentiment}, \textbf{climate\_fever},
            \textbf{medical\_questions\_pairs}, tweet\_eval-stance\_atheism, superglue-cb, dbpedia\_14, wiki\_qa, emo, yelp\_polarity, ethos-religion, \textbf{financial\_phrasebank},
            tab\_fact, anli, ethos-race \\
        \midrule
            Setting: QA Meta-train \\
            biomrc, boolq, freebase\_qa, hotpot\_qa, kilt\_hotpotqa, kilt\_nq, kilt\_trex, kilt\_zsre, lama-conceptnet, lama-google\_re, lama-squad, lama-trex, mc\_taco, numer\_sense, quoref, ropes, search\_qa, squad-no\_context, squad-with\_context, superglue-multirc, superglue-record, tweet\_qa, web\_questions, unifiedqa:squad2, unifiedqa:natural\_questions\_with\_dpr\_para, unifiedqa:race\_string, unifiedqa:squad1\_1, unifiedqa:drop, unifiedqa:newsqa, unifiedqa:narrativeqa, unifiedqa:winogrande\_xl, unifiedqa:social\_iqa, unifiedqa:quoref, unifiedqa:physical\_iqa, unifiedqa:ropes, unifiedqa:commonsenseqa, unifiedqa:boolq \\
        \midrule
            Setting: Non-QA Meta-train \\
            hate\_speech\_offensive, google\_wellformed\_query, circa, glue-sst2, scitail, emo, ag\_news, art, paws, kilt\_ay2, glue-qnli, ade\_corpus\_v2-classification, hatexplain, emotion, glue-qqp, kilt\_fever, dbpedia\_14, glue-mnli, discovery, gigaword, amazon\_polarity, tab\_fact, tweet\_eval-emoji, tweet\_eval-offensive, tweet\_eval-sentiment, imdb, liar, anli, wikisql, xsum, yahoo\_answers\_topics, yelp\_polarity, yelp\_review\_full\\ 
        \midrule
            Setting: QA Target \\
            ai2\_arc, codah, cosmos\_qa, dream, hellaswag, openbookqa, qasc, quail, quarel, quartz-no\_knowledge, quartz-with\_knowledge, sciq, superglue-copa, swag, wino\_grande, wiqa, unifiedqa:qasc, unifiedqa:qasc\_with\_ir, unifiedqa:openbookqa, unifiedqa:openbookqa\_with\_ir, \textbf{unifiedqa:mctest}, unifiedqa:ai2\_science\_middle\\
        \midrule
            Setting: Non-NLI Meta-train \\
            ade\_corpus\_v2-classification, ag\_news, amazon\_polarity, circa, climate\_fever, dbpedia\_14, discovery, emo, emotion, ethos-directed\_vs\_generalized,
            ethos-disability, ethos-gender, ethos-national\_origin, ethos-race, ethos-religion, ethos-sexual\_orientation, financial\_phrasebank, glue-cola, glue-mrpc,
            glue-qqp, glue-sst2, google\_wellformed\_query, hate\_speech18, hate\_speech\_offensive, hatexplain, health\_fact, imdb, kilt\_fever, liar, \\
            medical\_questions\_pairs, onestop\_english, paws, poem\_sentiment, rotten\_tomatoes, sick, sms\_spam, superglue-wic, superglue-wsc, tab\_fact,
            trec, trec-finegrained, tweet\_eval-emoji, tweet\_eval-emotion, tweet\_eval-hate, tweet\_eval-irony, tweet\_eval-offensive, tweet\_eval-sentiment,
            tweet\_eval-stance\_abortion, tweet\_eval-stance\_atheism, tweet\_eval-stance\_climate, tweet\_eval-stance\_feminist, tweet\_eval-stance\_hillary, wiki\_qa, yahoo\_answers\_topics, yelp\_polarity
            \\
            Setting: NLI Target \\
            anli, glue-mnli, glue-qnli, glue-rte, glue-wnli, \textbf{scitail}, sick, superglue-cb
            \\
        \midrule
            Setting: Non-Paraphrase Meta-train \\
            ade\_corpus\_v2-classification, ag\_news, amazon\_polarity, anli, circa, climate\_fever, dbpedia\_14, discovery, emo, emotion, ethos-directed\_vs\_generalized,
            ethos-disability, ethos-gender, ethos-national\_origin, ethos-race, ethos-religion, ethos-sexual\_orientation, financial\_phrasebank, glue-cola, glue-mnli, glue-qnli, 
            glue-rte, glue-sst2, glue-wnli, google\_wellformed\_query, hate\_speech18, hate\_speech\_offensive, hatexplain, health\_fact, imdb, kilt\_fever, liar, onestop\_english, 
            poem\_sentiment, rotten\_tomatoes, scitail, sick, sms\_spam, superglue-cb, superglue-rte, superglue-wic, superglue-wsc, tab\_fact, trec, trec-finegrained, tweet\_eval-emoji, 
            tweet\_eval-emotion, tweet\_eval-hate, tweet\_eval-irony, tweet\_eval-offensive, tweet\_eval-sentiment, tweet\_eval-stance\_abortion, tweet\_eval-stance\_atheism, 
            tweet\_eval-stance\_climate, tweet\_eval-stance\_feminist, tweet\_eval-stance\_hillary, wiki\_qa, yahoo\_answers\_topics, yelp\_polarity
            \\
        \midrule
            Setting: Non-Paraphrase Target \\
            Target: glue-mrpc, glue-qqp, \textbf{medical\_questions\_pairs}, paws
            \\
        \midrule
            Setting: \main\ Diverse Meta-train \\
            glue-mnli, glue-qqp, glue-sst2, hate\_speech\_offensive, kilt\_hotpotqa, kilt\_zsre, lama-trex, race-high, scitail, tweet\_eval-offensive, wino\_grande, yahoo\_answers\_topics, yelp\_review\_full \\
        \midrule
            Setting: \main\ No Diverse Meta-train \\
            ag\_news, amazon\_polarity, dbpedia\_14, emo, emotion, glue-sst2, imdb, tweet\_eval-emoji, tweet\_eval-offensive, tweet\_eval-sentiment, yahoo\_answers\_topics, yelp\_polarity, yelp\_review\_full \\
        \midrule
            Setting: \main\ Instructions Meta-train \\
            ag\_news, amazon\_polarity, anli, art, circa, cosmos\_qa, dbpedia\_14, discovery, emo, emotion, freebase\_qa, gigaword, google\_wellformed\_query, hellaswag, imdb, liar, paws, piqa, quail, quoref, ropes, sciq, scitail, social\_i\_qa, swag, tab\_fact, wiki\_qa, wiqa, xsum, yahoo\_answers\_topics, yelp\_polarity, yelp\_review\_full \\
        \midrule
            Setting: \main\ Instructions Target \\
            ai2\_arc, \textbf{climate\_fever}, codah, commonsense\_qa, dream, \textbf{financial\_phrasebank}, \textbf{medical\_questions\_pairs}, openbookqa, \textbf{poem\_sentiment}, qasc, quarel, sick \\
        \bottomrule
    \end{tabular}
    \caption{Full datasets for all settings.
    The first 10 rows are for main settings described in Section~\ref{subsec:dataset}; the last four rows are settings used for ablations in Section~\ref{subsec:ablations}.
    Splits and dataname names consistent to those in \citet{ye2021crossfit} and \citet{khashabi2020unifiedqa}.
    \textbf{Bold} indicates the test dataset with no overlap in domain with meta-training tasks. A prefix \texttt{unifiedqa:} indicates that the dataset taken is from \textsc{UnifiedQA}; otherwise, from \textsc{CrossFit}.
    References for all datasets are provided in Table~\ref{tab:full-datasets-citations}.
    }\label{tab:full-datasets}
\end{table*}

\begin{table*}[t]
    \centering \footnotesize
    \begin{tabular}{p{\textwidth}}
        \toprule
            ade\_corpus\_v2-classification~\citep{GURULINGAPPA2012885}, ade\_corpus\_v2-dosage~\citep{GURULINGAPPA2012885}, ag\_news~\href{http://groups.di.unipi.it/~gulli/AG_corpus_of_news_articles.html}{Gulli (link)}, ai2\_arc~\citep{Clark2018ThinkYH}, amazon\_polarity~\citep{McAuley2013HiddenFA}, anli~\citep{nie-etal-2020-adversarial}, art~\citep{bhagavatula2020abductive}, biomrc~\citep{pappas-etal-2020-biomrc}, blimp-anaphor\_number\_agreement~\citep{warstadt2019blimp}, blimp-ellipsis\_n\_bar\_2~\citep{warstadt2019blimp}, blimp-sentential\_negation\_npi\_licensor\_present~\citep{warstadt2019blimp}, blimp-sentential\_negation\_npi\_scope~\citep{warstadt2019blimp}, boolq~\citep{clark-etal-2019-boolq}, circa~\citep{louis-etal-2020-id}, climate\_fever~\citep{Diggelmann2020CLIMATEFEVERAD}, codah~\citep{chen-etal-2019-codah}, commonsense\_qa~\citep{talmor-etal-2019-commonsenseqa}, cosmos\_qa~\citep{huang-etal-2019-cosmos}, crows\_pairs~\citep{nangia-etal-2020-crows}, dbpedia\_14~\citep{Lehmann2015DBpediaA}, discovery~\citep{sileo-etal-2019-mining}, dream~\citep{sun-etal-2019-dream}, emo~\citep{chatterjee-etal-2019-semeval}, emotion~\citep{saravia-etal-2018-carer}, ethos-directed\_vs\_generalized~\citep{Mollas2020ETHOSAO}, ethos-disability~\citep{Mollas2020ETHOSAO}, ethos-gender~\citep{Mollas2020ETHOSAO}, ethos-national\_origin~\citep{Mollas2020ETHOSAO}, ethos-race~\citep{Mollas2020ETHOSAO}, ethos-religion~\citep{Mollas2020ETHOSAO}, ethos-sexual\_orientation~\citep{Mollas2020ETHOSAO}, financial\_phrasebank~\citep{financial-phrasebank}, freebase\_qa~\citep{jiang-etal-2019-freebaseqa}, gigaword~\citep{napoles-etal-2012-annotated}, glue-cola~\citep{warstadt-etal-2019-neural}, glue-mnli~\citep{williams-etal-2018-broad}, glue-mrpc~\citep{dolan-brockett-2005-automatically}, glue-qnli~\citep{rajpurkar-etal-2016-squad}, glue-qqp~(\url{data.quora.com/First-Quora-Dataset-Release-Question-Pairs}), glue-rte~\begin{tabular}[c]{@{}l@{}}\citep{dagan2005pascal, bar2006second}\citep{giampiccolo2007third, bentivogli2009fifth}\end{tabular}, glue-sst2~\citep{socher-etal-2013-recursive}, glue-wnli~\citep{levesque2012winograd}, google\_wellformed\_query~\citep{faruqui-das-2018-identifying}, hate\_speech18~\citep{gibert2018hate}, hate\_speech\_offensive~\citep{hateoffensive}, hatexplain~\citep{mathew2020hatexplain}, health\_fact~\citep{kotonya-toni-2020-explainable-automated}, hellaswag~\citep{zellers-etal-2019-hellaswag}, hotpot\_qa~\citep{yang-etal-2018-hotpotqa}, imdb~\citep{maas-etal-2011-learning}, kilt\_ay2~\citep{hoffart-etal-2011-robust}, kilt\_fever~\citep{thorne-etal-2018-fever}, kilt\_hotpotqa~\citep{yang-etal-2018-hotpotqa}, kilt\_nq~\citep{kwiatkowski-etal-2019-natural}, kilt\_trex~\citep{elsahar-etal-2018-rex}, kilt\_zsre~\citep{levy-etal-2017-zero}, lama-conceptnet~\citep{petroni-etal-2019-language,petroni2020how}, lama-google\_re~\citep{petroni-etal-2019-language,petroni2020how}, lama-squad~\citep{petroni-etal-2019-language,petroni2020how}, lama-trex~\citep{petroni-etal-2019-language,petroni2020how}, liar~\citep{wang-2017-liar}, mc\_taco~\citep{zhou-etal-2019-going}, medical\_questions\_pairs~\citep{medical-qqp}, numer\_sense~\citep{lin-etal-2020-birds}, onestop\_english~\citep{vajjala-lucic-2018-onestopenglish}, openbookqa~\citep{mihaylov-etal-2018-suit}, paws~\citep{zhang-etal-2019-paws}, piqa~\citep{bisk2019piqa}, poem\_sentiment~\citep{sheng-uthus-2020-investigating}, proto\_qa~\citep{boratko-etal-2020-protoqa}, qa\_srl~\citep{he-etal-2015-question}, qasc~\citep{Khot_Clark_Guerquin_Jansen_Sabharwal_2020}, quail~\citep{Rogers_Kovaleva_Downey_Rumshisky_2020}, quarel~\citep{Tafjord_Clark_Gardner_Yih_Sabharwal_2019}, quartz-no\_knowledge~\citep{tafjord-etal-2019-quartz}, quartz-with\_knowledge~\citep{tafjord-etal-2019-quartz}, quoref~\citep{dasigi-etal-2019-quoref}, race-high~\citep{lai-etal-2017-race}, race-middle~\citep{lai-etal-2017-race}, ropes~\citep{lin-etal-2019-reasoning}, rotten\_tomatoes~\citep{pang-lee-2005-seeing}, sciq~\citep{welbl-etal-2017-crowdsourcing}, scitail~\citep{scitail}, search\_qa~\citep{Dunn2017SearchQAAN}, sick~\citep{marelli-etal-2014-sick}, sms\_spam~\citep{sms_spam}, social\_i\_qa~\citep{sap-etal-2019-social}, spider~\citep{yu-etal-2018-spider}, squad-no\_context~\citep{rajpurkar-etal-2016-squad}, squad-with\_context~\citep{rajpurkar-etal-2016-squad}, superglue-cb~\citep{Marneffe_Simons_Tonhauser_2019}, superglue-copa~\citep{gordon-etal-2012-semeval}, superglue-multirc~\citep{khashabi-etal-2018-looking}, superglue-record~\citep{Zhang2018ReCoRDBT}, superglue-rte~\begin{tabular}[c]{@{}l@{}}\citep{dagan2005pascal, bar2006second}\citep{giampiccolo2007third, bentivogli2009fifth}\end{tabular}, superglue-wic~\citep{pilehvar-camacho-collados-2019-wic}, superglue-wsc~\citep{levesque2012winograd}, swag~\citep{zellers-etal-2018-swag}, tab\_fact~\citep{Chen2020TabFact}, trec~\citep{li-roth-2002-learning,hovy-etal-2001-toward}, trec-finegrained~\citep{li-roth-2002-learning,hovy-etal-2001-toward}, tweet\_eval-emoji~\citep{barbieri-etal-2020-tweeteval}, tweet\_eval-emotion~\citep{barbieri-etal-2020-tweeteval}, tweet\_eval-hate~\citep{barbieri-etal-2020-tweeteval}, tweet\_eval-irony~\citep{barbieri-etal-2020-tweeteval}, tweet\_eval-offensive~\citep{barbieri-etal-2020-tweeteval}, tweet\_eval-sentiment~\citep{barbieri-etal-2020-tweeteval}, tweet\_eval-stance\_abortion~\citep{barbieri-etal-2020-tweeteval}, tweet\_eval-stance\_atheism~\citep{barbieri-etal-2020-tweeteval}, tweet\_eval-stance\_climate~\citep{barbieri-etal-2020-tweeteval}, tweet\_eval-stance\_feminist~\citep{barbieri-etal-2020-tweeteval}, tweet\_eval-stance\_hillary~\citep{barbieri-etal-2020-tweeteval}, tweet\_qa~\citep{xiong-etal-2019-tweetqa}, unifiedqa:ai2\_science\_middle~(\url{data.allenai.org/ai2-science-questions}), unifiedqa:boolq~\citep{clark-etal-2019-boolq}, unifiedqa:commonsenseqa~\citep{talmor-etal-2019-commonsenseqa}, unifiedqa:drop~\citep{dua-etal-2019-drop}, unifiedqa:mctest~\citep{richardson-etal-2013-mctest}, unifiedqa:narrativeqa~\citep{kocisky-etal-2018-narrativeqa}, unifiedqa:natural\_questions~\citep{kwiatkowski-etal-2019-natural}, unifiedqa:newsqa~\citep{trischler-etal-2017-newsqa}, unifiedqa:openbookqa~\citep{mihaylov-etal-2018-suit}, unifiedqa:physical\_iqa~\citep{bisk2019piqa}, unifiedqa:qasc~\citep{khot2019qasc}, unifiedqa:quoref~\citep{dasigi-etal-2019-quoref}, unifiedqa:race\_string~\citep{lai-etal-2017-race}, unifiedqa:ropes~\citep{lin-etal-2019-reasoning}, unifiedqa:social\_iqa~\citep{sap2019socialiqa}, unifiedqa:squad1\_1~\citep{rajpurkar-etal-2016-squad}, unifiedqa:squad2~\citep{rajpurkar-etal-2018-know}, unifiedqa:winogrande\_xl~\citep{sakaguchi2019winogrande}, web\_questions~\citep{berant-etal-2013-semantic}, wiki\_qa~\citep{yang-etal-2015-wikiqa}, wikisql~\citep{zhongSeq2SQL2017}, wino\_grande~\citep{Sakaguchi_Le_Bras_Bhagavatula_Choi_2020}, wiqa~\citep{tandon-etal-2019-wiqa}, xsum~\citep{narayan-etal-2018-dont}, yahoo\_answers\_topics~\href{https://webscope.sandbox.yahoo.com/catalog.php?datatype=l}{(link)}, yelp\_polarity~\citep{zhang2015character}, yelp\_review\_full~\citep{zhang2015character} \\
        \bottomrule
    \end{tabular}
    \caption{
    References for 142 datasets used in the paper.
    A prefix \texttt{unifiedqa:} indicates that the dataset taken is from \textsc{UnifiedQA}; otherwise, from \textsc{CrossFit}.
    }\label{tab:full-datasets-citations}
\end{table*}

\section{Potential Risks}
\ours\ is based on the large language model that is pretrained on a web corpus, which potentially includes harmful and biased context, despite the original authors' best efforts to mine the text.
There are also potential risks in privacy and security---for instance, \citet{carlini2021extracting} reported that it is possible to design the attack algorithm to extract a substantial amount of training data.
We thus highlight that \ours\ should be considered as a research prototype rather than a deployable system to real users, and continuing efforts are needed to reduce potential risks of the model.

\end{document}